\documentclass{svjour3}                     
\smartqed  

\usepackage{graphicx}
\usepackage{mathptmx}
\usepackage{cite}
\usepackage{booktabs}
\usepackage{multirow}
\usepackage{array}
\usepackage{enumerate}
\usepackage{xcolor}
\definecolor{green}{rgb}{0,0.5,0}
\definecolor{red}{rgb}{0.8,0,0}
\definecolor{blue}{rgb}{0,0,0.8}
\definecolor{orange}{rgb}{0.85,0.35,0}
\usepackage{natbib}
\usepackage{pdflscape}
\usepackage{url}
\newcolumntype{x}[1]{%
>{\centering\hspace{0pt}}p{#1}}%

\journalname{Artificial Intelligence Review}

\begin{document}

\title{Artificial Intelligence in the Creative Industries: A Review}
\author{Nantheera Anantrasirichai \and David Bull} 

\institute{Nantheera Anantrasirichai and David Bull \at 
Bristol Vision Institute, University of Bristol, UK \\ \email{n.anantrasirichai@bristol.ac.uk; dave.bull@bristol.ac.uk}}

\date{Accepted for publication in Artificial Intelligence Review (AIRE), 19 June 2021}

\maketitle

\begin{abstract}
    This paper reviews the current state of the art in Artificial Intelligence (AI) technologies and applications in the context of the creative industries. A brief background of AI, and specifically Machine Learning (ML) algorithms, is provided including Convolutional Neural Networks (CNNs), Generative Adversarial Networks (GANs), Recurrent Neural Networks (RNNs) and Deep Reinforcement Learning (DRL). We {categorize} creative applications into five groups, related to how AI technologies are used: i) content creation, ii) information analysis, iii) content enhancement and post production workflows, iv) information extraction and enhancement, and v) data compression.  We critically examine the successes and limitations of this rapidly advancing technology in each of these areas. We further differentiate between the use of AI as a creative tool and its potential as a creator in its own right.  We foresee that, in the near future, ML-based AI will be adopted widely as a tool or collaborative assistant for creativity.  In contrast, we observe that the successes of ML in domains with fewer constraints, where AI is the `creator', remain modest. The potential of AI (or its developers) to win awards for its original creations in competition with human creatives is also limited, based on contemporary technologies.  We therefore conclude that, in the context of creative industries, maximum benefit from AI will be derived where its focus is human-centric -- where it is designed to augment, rather than replace, human creativity.
    

\end{abstract}

\section{Introduction}
\label{sec:intro}

The aim of new technologies is normally to make a specific process easier, more accurate, faster or cheaper. In some cases they also enable us to perform tasks or create things that were previously impossible. Over recent years, one of the most rapidly advancing scientific techniques for practical purposes has been Artificial Intelligence (AI). AI techniques enable machines to perform tasks that typically require some degree of human-like intelligence. With recent developments in high-performance computing and increased data storage capacities, AI technologies have been empowered and are increasingly being adopted across numerous applications, ranging from simple daily tasks, intelligent assistants and finance to highly specific command, control operations and national security.  AI can, for example, help smart devices or computers to understand text and read it out loud, hear voices and respond, view images and recognize objects in them, and even predict what may happen next after a series of events. At higher levels, AI has been used to analyze human and social activity by observing their convocation and actions. It has also been used to understand socially relevant problems such as homelessness and to predict natural events. 
AI has been recognized by governments across the world to have potential as a major driver of economic growth and social progress \citep{Executive:Preparing:2016, Hall:Growing:2018}. This potential, however, does not come without concerns over the wider social impact of AI technologies which must be taken into account when designing and deploying these tools. 

Processes associated with the creative sector demand  significantly different levels of innovation and  skill sets compared to routine behaviours. {While} AI accomplishments rely heavily on conformity of data, creativity often exploits the human imagination to drive original ideas which may not follow general rules. Basically, creatives have a lifetime of experiences to build on, enabling them to think `outside of the box' and ask `What if' questions that cannot readily be addressed by constrained learning systems.  

There have been many studies over several decades  into the possibility of applying AI in the creative sector. One of the limitations in the past was the readiness of the technology itself, and another was the belief that AI could attempt to replicate human creative behaviour \citep{Rowe:Subject:1993}. A recent survey by Adobe\footnote{\url{https://www.pfeifferreport.com/wp-content/uploads/2018/11/Creativity_and_AI_Report_INT.pdf}}  revealed that three quarters of artists in the US, UK, Germany and Japan would consider using AI tools as assistants, in areas such as image search, editing, and other `non-creative' tasks. This indicates a general acceptance of AI as a tool across the community and reflects a general awareness of the state of the art, since  most AI technologies have been developed to operate in closed domains where they can assist and support humans rather than replace them. Better collaboration between humans and AI technologies can thus maximize the benefits of the synergy. All that said, the first painting created solely by AI was auctioned for \$432,500 in 2018\footnote{https://edition.cnn.com/style/article/obvious-ai-art-christies-auction-smart-creativity/index.html}. 

{Applications of AI in the creative industries have dramatically increased in the last five years. Based on analysis of data from arXiv\footnote{https://arxiv.org/} and Gateway to Research\footnote{https://gtr.ukri.org/}, \citet{Davies:Art:2020} revealed that the growth rate of research publications on AI (relevant to the creative industries) exceeds  500\% in many countries (in Taiwan the growth rate is 1,490\%), and the most of these publications relate to image-based data. Analysis on company usage from the Crunchbase database\footnote{https://www.crunchbase.com/} indicates that AI is used more in games and for immersive applications, advertising and marketing, than in other creative applications. \citet{Briot:AI:2019} recently  reviewed  AI in the current media and creative industries across three areas: creation, production and consumption. They provide  details of AI/ML-based research and development, as well as  emerging challenges and trends.}

{In this paper, we review how AI and its technologies are, or could be, used in applications relevant to creative industries. We first provide an overview of AI and current technologies (Section 2), followed by a selection of creative domain applications (Section \ref{sec:exiting}). We group these into subsections\footnote{While we hope that this categorization is helpful, it should be noted that several of the applications described could fit into, or span, multiple categories.} covering: i) content creation: where AI is employed to generate original work, ii) information analysis: where statistics of data are used to improve productivity,  iii) content enhancement and post production workflows: used to improve quality of creative work, iv) information extraction and enhancement: where AI assists in interpretation, clarifies semantic meaning, and creates new ways to exhibit hidden information, and v) data compression: where AI helps reduce the size of the data while preserving its quality. Finally we discuss challenges and the future potential of AI associated with the  creative industries in Section \ref{sec:discussion}. }

{
\section{An introduction to Artificial Intelligence}}
\label{sec:overview}

{Artificial intelligence (AI) embodies a set of codes}, techniques, algorithms and data that enables a computer system to develop and emulate human-like behaviour and hence make decisions similar to (or in some cases, better than) humans \citep{Russell:Artificial:2020}. When a machine  exhibits full human intelligence, it is often referred to as `general AI' or `strong AI' \citep{Bostrom:Superintelligence:2014}. However, currently reported technologies are normally restricted to operation in a limited domain to work on specific tasks. This is called `narrow AI' or `weak AI'. In the past, most AI technologies were model-driven; where the nature of the application is studied and a model is mathematically formed to describe it.  Statistical learning is also data-dependent, but relies on rule-based programming \citep{James:Statistical:2013}. Previous generations of AI (mid-1950s until the late 1980s \citep{Haugeland:Artificial:1985}) were based on symbolic AI, following the assumption that {humans use symbols} to represent things and problems. Symbolic AI is intended to produce general, human-like intelligence in a machine \citep{Honavar:Symbolic:1995}, whereas most modern research is directed at specific sub-problems.

\subsection{Machine Learning, Neurons and Artificial Neural Networks}
\label{ssec:ml}
The main class of algorithms in use today are based on machine learning (ML), which is data-driven. ML employs computational methods to `learn' information directly from large amounts of example data without relying on a predetermined equation or model \citep{Mitchell:ML:1997}. These algorithms adaptively converge to an optimum solution and generally improve their performance as the number of samples available for learning increases. {Several types of learning algorithms exist}, including supervised learning, unsupervised learning and reinforcement learning. Supervised learning algorithms build a mathematical model from a set of data that contains both the inputs and the desired outputs (each output usually representing a classification of the associated input vector), {while} unsupervised learning algorithms model the problems on unlabeled data.  {Self-supervised learning is a form of unsupervised learning where the data provide the  measurable structure to build a loss function.
Semi-supervised learning employs a  limited set of labeled data to label, usually a larger amount of, unlabeled data. Then both datasets are combined to create a new model. } Reinforcement learning methods learn from trial and error and are effectively self-supervised \citep{Russell:Artificial:2020}.

\begin{figure*}[t!]
	\centering
	\vspace{5mm}
      		 \includegraphics[width=\textwidth]{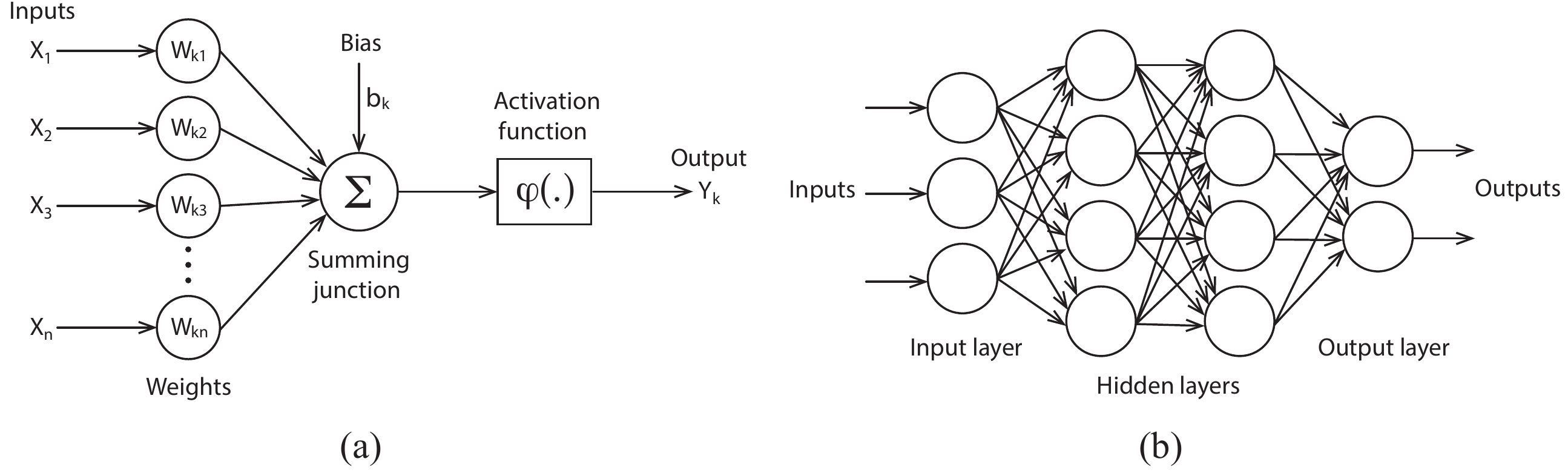}  
					\caption{\small (a) Basic neural network unit by MuCulloch and Pitts. (b) Basic multi-layer perceptron (MLP).}
    \label{fig:ANN}
\end{figure*}

Modern ML methods have their roots in the early computational model of a neuron proposed by Warren MuCulloch (neuroscientist) and Walter Pitts (logician) in \citeyear{McCulloch:logical:1943}. This is shown in Fig. \ref{fig:ANN} (a). In their model, the artificial neuron receives one or more inputs, where each input is independently weighted. The neuron sums these weighted inputs and the result is passed through  a non-linear function known as an activation function, representing the neuron's action potential which is then transmitted along its axon to other neurons. The multi-layer perceptron (MLP) is a basic form of Artificial Neural Network (ANN) that gained popularity in the 1980s. This connects its neural units in a multi-layered  (typically one input layer, one hidden layer and one output layer) architecture (Fig. \ref{fig:ANN} (b)). These neural layers are generally fully connected to adjacent layers, (i.e., each neuron in one layer is connected to all neurons in the next layer). The disadvantage of this approach is that the total number of parameters can be very large and this can make them prone to overfitting data.

For training, the MLP (and most supervised ANNs) utilizes error backpropagation to compute the gradient of a loss function. This loss function maps the event values from multiple inputs into one real number to represent the cost of that event. The goal of the training process is therefore to minimize the loss function over multiple presentations of the input dataset. The backpropagation algorithm was originally introduced in the 1970s, but peaked in popularity after 1986, when \citeauthor{Rumelhart:Learning:1986} described several neural networks where backpropagation worked far faster than earlier approaches, making ANNs applicable to practical problems. 

\subsection{An Introduction to Deep Neural Networks}

Deep learning is a subset of ML that employs deep artificial neural networks (DNNs). The word `deep' means that there are multiple hidden layers of neuron collections that have learnable weights and biases. When the data being processed occupies multiple dimensions (images for example), Convolutional Neural Networks (CNNs) are often employed. CNNs are (loosely) a biologically-inspired architecture and their results are tiled so that they overlap to obtain a better representation of the original inputs. 

\begin{figure*}[t!]
	\centering
	\vspace{5mm}
      		 \includegraphics[width=\textwidth]{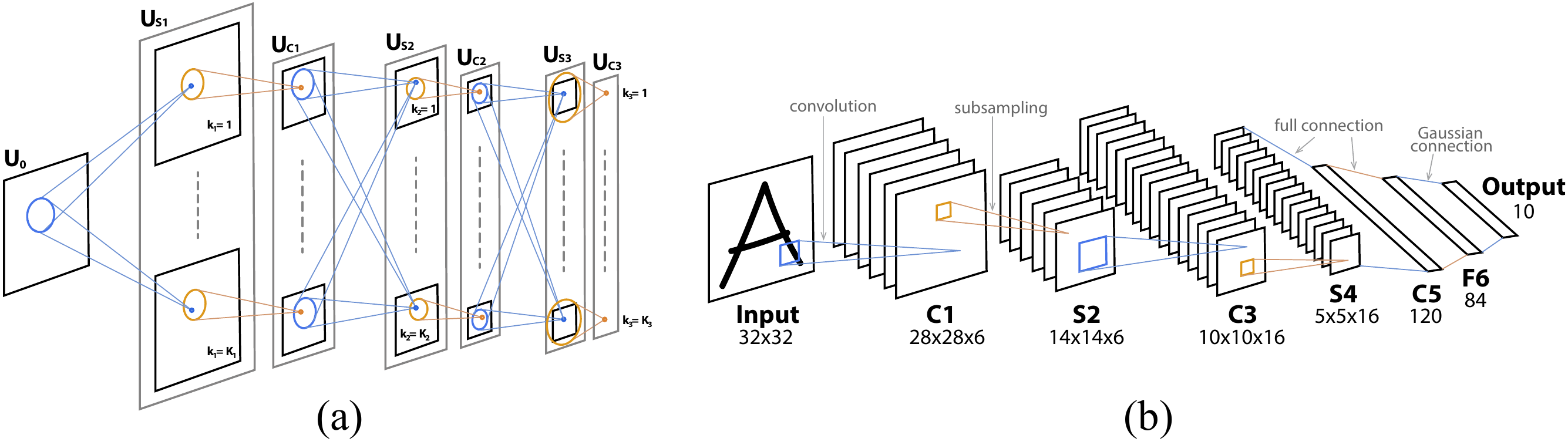}  
					\caption{\small (a)  Neocognitron \citep{Fukushima:Neocognitron:1980}, where U$_s$ and U$_c$ learn simple and complex features, respectively.  (b) LeNet5 \citep{Lecun:Gradient:1998}, consisting of two sets of convolutional and average pooling layers, followed by a flattening convolutional layer, then two fully-connected layers and finally a softmax classifier.}
    \label{fig:earlyDNN}
\end{figure*}

The first CNN was designed by \citet{Fukushima:Neocognitron:1980} as a tool for visual pattern recognition  (Fig. \ref{fig:earlyDNN} (a)). This so called Neocognitron was a hierarchical architecture with multiple convolutional and pooling layers. In 1989, \citeauthor{LeCun:Backpropagation:1989} applied the standard backpropagation algorithm  to a deep neural network with the purpose of recognizing handwritten ZIP codes. At that time, it took 3 days to train the network. In 1998, \citeauthor{Lecun:Gradient:1998} proposed LeNet5 (Fig. \ref{fig:earlyDNN} (b)), one of the earliest CNNs which could outperform other models for handwritten character recognition. The deep learning breakthrough occurred in the 2000s driven by the availability of graphics processing units (GPUs) that could dramatically accelerate training. Since around 2012, CNNs have represented the state of the art for complex problems such as image classification and recognition, having won  several major international competitions. 

A CNN creates its filters’ values based on the task at hand. Generally, the CNN learns to detect edges from the raw pixels in the first layer, then uses those edges to detect simple shapes in the next layer, and so on building complexity through subsequent layers. The higher layers produce high-level features with more semantically relevant meaning. This means that the algorithms can exploit both low-level features and a higher-level understanding of what the data represent. Deep learning has therefore emerged as a powerful tool to find patterns, analyze information, and to predict future events. The number of layers in a deep network is unlimited but most current networks contain between 10 and 100 layers. 

In 2014, \citeauthor{Goodfellow:GAN:2014} proposed an alternative form of architecture referred to as a Generative Adversarial Network (GAN). GANs consist of 2 AI competing modules where the first creates images (the generator) and the second (the discriminator) checks whether the received image is real or created from the first module.  This competition results in the final picture  being very similar to the real image. Because of their performance in reducing deceptive results, GAN technologies have become very popular and have been applied to numerous applications, including those related to creative practice.


While many types of machine learning algorithms exist, because of their prominence and performance, in this paper we place emphasis on deep learning methods. We  will describe various applications relevant to the creative industries and critically review the methodologies that achieve, or have the potential to achieve, good performance. 

\subsection{Current AI technologies}
\label{sec:methods}

This section presents state-of-the-art AI methods relevant to the creative industries. For those readers who prefer to focus on the applications, please refer to Section \ref{sec:exiting}.

\subsubsection{AI and the Need for Data}
\label{ssec:datasets}
An AI system effectively combines a computational architecture and a learning strategy with a data environment in which it learns. Training databases are thus a critical component in {optimizing} the performance of ML processes and hence a significant proportion of the value of an AI system resides in them. A well-designed training database with appropriate size and coverage can help significantly with model {generalization} and avoiding problems of {overfitting}.  

 In order to learn without being explicitly programmed, ML systems must be trained using data having statistics and characteristics typical of the particular application domain under consideration.  { This is true regardless of training methods (see Section \ref{ssec:ml}).}  Good  datasets typically contain large numbers of examples with a statistical distribution matched to this domain. This is crucial because it enables the network to estimate gradients in the data (error) domain that enables it to converge to an optimum solution, forming robust decision boundaries between its classes.  The network will then, after training,  be able to reliably match new unseen  information to the right answer when deployed.
 
 The reliability of training dataset labels is key in achieving high performance supervised deep learning. These datasets must comprise: i) data that are statistically similar to the inputs when the models are used in the real situations and ii) ground truth annotations that tell the machine what the desired outputs are. For example, in  segmentation applications, the dataset would comprise the images and the corresponding segmentation maps indicating homogeneous, or semantically meaningful regions in each image. Similarly for object recognition, the dataset would also include the original images while the ground truth would be the object categories, e.g., car, house, human, type of animals, etc. 

Some labeled datasets are freely available for public  use\footnote{https://en.wikipedia.org/wiki/List\_of\_datasets\_for\_machine-learning\_research, https://ieee-dataport.org/}, but these are limited, especially in certain applications where data are difficult to collect and label. One of the largest, ImageNet, contains over 14 million images labeled into 22,000 classes. {Care must be taken when collecting or using} data to avoid imbalance and bias --  skewed class distributions where the majority of data instances belong to a small number of classes  with other  classes being sparsely populated.  For instance, in colorization, blue may appear more often as it is a color of sky, {while} pink flowers are much rarer. This imbalance causes ML algorithms to develop a bias towards classes with a greater number of instances;  hence they preferentially predict majority class data. Features of minority classes are treated as noise and are often ignored. 

Numerous approaches have been introduced to create balanced distributions and these can be divided into two major groups: modification of the learning algorithm, and data manipulation techniques \citep{He:Learning:2009}.  \citet{Zhang:colorful:2016} solve the class-imbalance problem by re-weighting the loss of each pixel at train time based on the pixel color rarity. Recently, \citet{Lehtinen:noise2noise:2018} have introduced an innovative approach to learning via their Noise2Noise network  which demonstrates that it is possible to train a network without clean data if the corrupted data complies with certain statistical assumptions. However, this technique needs further testing and refinement to cope with  real-world noisy data. Typical data manipulation techniques include downsampling majority classes, oversampling minority classes, or both. 
Two primary techniques are used to expand, adjust and rebalance the number of samples in the dataset and, in turn,  to improve ML performance and model {generalization}: data augmentation and data synthesis. These are discussed further below.

{i) Data augmentation}

 Data augmentation techniques are frequently used to increase the volume and diversity of a training dataset without the need to collect new data. Instead,   existing data are used to generate more samples, through transformations such as cropping, flipping, translating,  rotating and scaling \citep{Krizhevsky:ImageNet:2012, Anantrasirichai:Application:2018}. This can assist by increasing the representation of  minority classes and also help to avoid overfitting, which occurs when a model memorizes the full dataset instead of only learning the main concepts which underlie the problem. GANs (see Section \ref{ssec:GANs}) have recently been employed with success to enlarge training sets,  with the most popular  network currently being CycleGAN \citep{Zhu:Unpaired:2017}. The original CycleGAN mapped one input to only one output, causing inefficiencies when dataset diversity is required. \citet{Huang:AugGAN:2018} improved CycleGAN with a structure-aware network to augment training data for vehicle detection. This slightly modified architecture is trained to transform contrast CT images (computed tomography scans) into non-contrast images \citep{Sandfort:augmentation:2019}. A CycleGAN-based technique has also been used for emotion classification, to amplify cases of extremely rare emotions such as disgust \citep{Zhu:Emotion:2018}. IBM Research introduced a Balancing GAN \citep{Mariani:BAGAN:2018}, where the model learns useful features from majority classes and uses these to generate images for minority classes that avoid features close to those of majority cases. An extensive survey of data augmentation techniques can be found in \citep{Shorten:survey:2019}. 

 {ii) Data synthesis}

 Scientific or parametric models can be exploited to generate synthetic data in those applications where it is difficult to collect real data, and where data augmentation techniques cannot increase variety in the dataset. Examples include signs of disease \citep{Alsaih:Machine:2017} and geological events that  rarely happen \citep{Anantrasirichai:deep:2019}. In the case of creative processes, problems are often ill-posed as ground truth data or ideal outputs are not available. Examples include post-production operations such as deblurring, denoising and contrast enhancement.  Synthetic data are often created by degrading the clean data. \citet{Su:Deep:2017} applied synthetic motion blur on sharp video frames to train the deblurring model. LLNet \citep{Lore:LLNet:2017}, enhances low-light images, and is trained using a  dataset generated with synthetic noise and intensity adjustment, {while} LLCNN \citep{Tao:LLCNN:2017} employs a gamma adjustment technique. 

\begin{figure*}[t!]
	\centering
	\vspace{5mm}
      		 \includegraphics[width=0.75\textwidth]{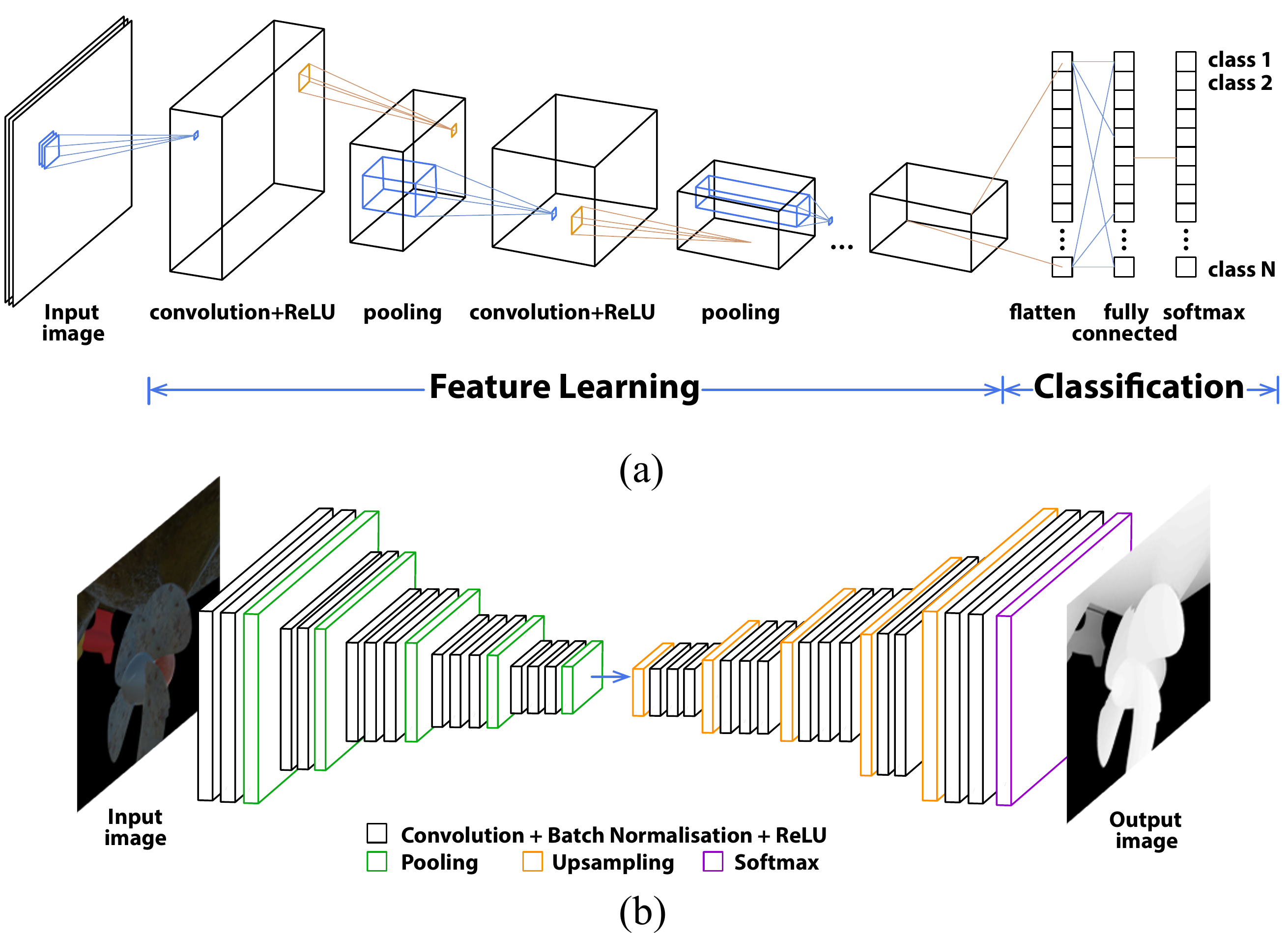}  
					\caption{\small CNN architectures for (a) object recognition adapted from{ $^{\ref{fn:CNN}}$}, (b) semantic segmentation{ $^{\ref{fn:semantic}}$}}
    \label{fig:CNNexample}
\end{figure*} 

\subsubsection{Convolutional Neural Networks (CNNs)}
\label{ssec:cnn}

{{i) Basic CNNs}}

Convolutional Neural Networks (CNNs) are a class of deep feed-forward ANN. They comprise a series of convolutional layers that are designed to take advantage of 2D structures, such as found in images. These employ locally connected layers that apply convolution operations between a predefined-size kernel and an internal signal; the output of each convolutional layer is the input signal modified by a convolution filter. The weights of the filter are adjusted according to a loss function that assesses the mismatch (during training) between the network output and the ground truth values or labels. Commonly used loss functions include $\ell_1$, $\ell_2$, SSIM \citep{Tao:LLCNN:2017} and perceptual loss \citep{Johnson:Perceptual:2016}). These errors are then backpropagated  through multiple forward and backward iterations and the filter weights adjusted based on estimated gradients of the local error surface.  This in turn drives what features are detected, associating them to the characteristics of the training data. The early layers in a CNN extract low-level features conceptually similar to visual basis functions found in the primary visual cortex \citep{Matsugu:Subject:2003}. 

The most common CNN architecture (Fig. \ref{fig:CNNexample} (a){\footnote{\label{fn:CNN}https://uk.mathworks.com/solutions/deep-learning/convolutional-neural-network.html}}) has the outputs from its convolution layers connected to a pooling layer, which combines the outputs of neuron clusters into a single neuron.  Subsequently, activation functions such as {\textit{tanh}} (the hyperbolic tangent) or {\textit{ReLU}} (Rectified Linear Unit) are applied to introduce non-linearity into the network \citep{Agostinelli:Learning:2015}. This structure is repeated with similar or different kernel sizes. As a result, the CNN learns to detect edges from the raw pixels in the first layer, then combines these to detect simple shapes in the next layer. The higher layers produce higher-level features, which have more semantic meaning. The last few layers represent the classification part of the network. These consist of fully connected layers (i.e. being connected to all the activation outputs in the previous layer) and a softmax layer, where the output class is modelled as a probability distribution - exponentially scaling the output between 0 and 1 (this is also referred to as a normalised exponential function). 

{
VGG \citep{Simonyan:VGG:2015} is one of the most common backbone networks, offering two depths: VGG-16 and VGG-19 with 16 and 19 layers respectively. The networks incorporate a series of convolution blocks (comprising convolutional layers, ReLU activations and a max-pooling layer), and the last three layers are fully connection with ReLU activations. VGG employs very small receptive fields (3$\times$3 with a stride of 1) allowing deeper architectures than the older networks.} DeepArt  \citep{Gatys:Neural:2016} employs a VGG-Network  without fully connected layers. It demonstrates that the higher layers in the VGG network can represent the content of an artwork. The pre-trained VGG network is widely used to provide a measure of perceptual loss (and style loss) during the training process of other networks  \citep{Johnson:Perceptual:2016}.

{{ii) CNNs with reconstruction}}

The basic structure of CNNs described in the previous section is sometimes called an `encoder'. This is because the network learns a representation of a set of data, which often has fewer parameters than the input. In other words, it compresses the input to produce a code or a latent-space representation. In contrast,
some architectures omit pooling layers in order to create dense features in an output with the same size as the input. 

Alternatively, the size of the feature map can be enlarged to that of the input via deconvolutional layers or transposed convolution layers (Fig. \ref{fig:CNNexample} (b){\footnote{\label{fn:semantic} https://uk.mathworks.com/solutions/image-video-processing/semantic-segmentation.html}}). 
{This structure is often referred to as a  `decoder' as it generates the output using the code produced by the encoder. Encoder-decoder architectures combine an encoder and a decoder.  Autoencoders are a special case of encoder-decoder models, where the input and output are the same size.} Encoder-decoder models are suitable for creative applications, such as style transfer \citep{Zhang:colorful:2016}, image restoration \citep{Nah:Deep:2017, Zhang:dncnn:2017, Yang:Proximal:2018}, contrast enhancement \citep{Lore:LLNet:2017, Tao:LLCNN:2017}, colorization \citep{Zhang:colorful:2016} and super-resolution \citep{Shi:Real:2016}. 

Some architectures also add skip connections or a bridge section \citep{Long:fully:2015} so that the local and global features, as well as semantics are connected and captured, providing improved pixel-wise accuracy. These techniques are widely used in object detection \citep{Anantrasirichai:DefectNET:2019} and object tracking \citep{Redmon:YOLOv3:2018}. 
{
U-Net \citep{Ronneberger:Unet:2015} is perhaps the most popular network of this kind, even though it was originally developed for biomedical image segmentation.  Its network consists of a contracting path (encoder) and an expansive path (decoder), giving it the u-shaped architecture.  The contracting path consists of the repeated application of two 3$\times$3 convolutions, followed by ReLU and a max-pooling layer. Each step in the expansive path consists of a transposed convolution layer for upsampling, followed by two sets of convolutional and ReLU layers, and concatenations with correspondingly-resolution features from the contracting path.
}

{{iii) Advanced CNNs}}

 Some architectures introduce modified convolution operations for specific applications. For example, dilated convolution \citep{Yu:Multi:2016}, also called atrous convolution, enlarges the receptive field, to support feature extraction locally and globally. {The dilated convolution is applied to the input with a defined spacing between the values in a kernel. For example,  a 3$\times$3 kernel with a dilation rate of 2 has the same receptive field as a 5$\times$5 kernel, but using 9 parameters.} This has been used for colorization by \citet{Zhang:colorful:2016} in the creative sector. ResNet is an architecture developed for residual learning, comprising several residual blocks \citep{He:ResNet:2016}. A single residual block has two convolution layers and a skip connection between the input and the output of the last convolution layer. This avoids the problem of vanishing gradients, enabling very deep CNN architectures. Residual learning has become an important part of the state of the art in many application, such as contrast enhancement \citep{Tao:LLCNN:2017}, colorization \citep{Huang:Densely:2017}, SR \citep{Zhang:image:2018,Dai:Second:2019}, object recognition \citep{He:ResNet:2016}, and denoising \citep{Zhang:dncnn:2017}. 

Traditional convolution operations are performed in a regular grid fashion, leading to limitations for some applications, where the object and its location are not in the regular grid. Deformable convolution \citep{Dai:Deformable:2017} has therefore been proposed to facilitate the region of support for the convolution operations to take on any shape, instead of just the traditional square shape. This has been used in object detection and SR \citep{Wang:SR:2019}. 3D deformable kernels have also been proposed for denoising video content, as they can better cope with large motions, producing cleaner and sharper sequences \citep{Xu:Deformable:2019}.

{
Capsule networks were developed to address some of the deficiencies with traditional CNNs \citep{Sabour:Dynamic:2017}. They are able to better model hierarchical relationships, where each neuron (referred to as a capsule) expresses the likelihood and properties of its features, e.g., orientation or size. This improves object recognition performance.  Capsule networks  have been extended to other applications that deal with complex data, including multi-label text classification \citep{Zhao:towards:2019}, slot filling and intent detection \citep{Zhang:Joint:2019}, polyphonic sound event detection \citep{Vesperini:Polyphonic:2019} and sign language recognition \citep{Jalal:American:2018}.
}

\subsubsection{Generative Adversarial Networks (GANs)}
\label{ssec:GANs}

\begin{figure*}[t!]
	\centering
	\vspace{5mm}
      		 \includegraphics[width=\textwidth]{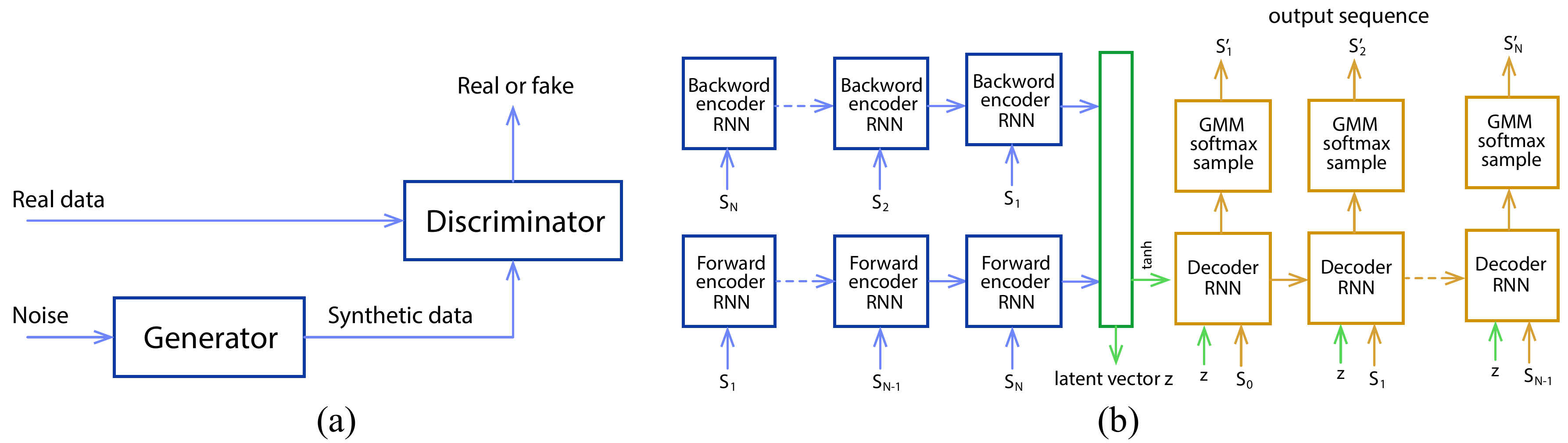}  
					\caption{\small Architectures of (a) GAN, (b) RNN for drawing sketches \citep{Ha:Neural:2017}}
    \label{fig:GANdiagrams}
\end{figure*}

The Generative Adversarial Network (GAN) is a recent algorithmic {innovation that employs} two neural networks: generative and  discriminative. The GAN pits one against the other in order to generate new, synthetic instances of data that can pass for real data. The general GAN architecture is shown in Fig. \ref{fig:GANdiagrams} (a). It can be observed that the generative network generates new candidates to increase the error rate of the discriminative network until the discriminative network cannot tell whether these candidates are real or synthesized. The generator is typically a deconvolutional neural network, and the discriminator is a CNN. Recent successful applications of GANs include SR \citep{Ledig:Photo:2017}, inpainting \citep{Yu:Freeform:2019}, contrast enhancement \citep{Kuang:single:2019} and compression \citep{Ma:Perceptually:2019}.

GANs have a reputation of being difficult to train since the two models are trained simultaneously to find a Nash equilibrium but with each model updating its cost (or error) independently. Failures often occur when the discriminator cannot feedback information that is good enough for the generator to make progress, leading to vanishing gradients. Wasserstein loss is designed to prevent this \citep{Frogner:Learning:2015, Arjovsky:Wasserstein:2017}. A specific condition or characteristic, such as a label associated with an image, rather than a generic sample from an unknown noise distribution can be included in the generative model, creating what is referred to as a conditional GAN (cGAN) \citep{Mirza:Conditional:2014}. This improved GAN has been used in several applications, including pix2pix\citep{Isola:Image:2017} and for deblurring \citep{Kupyn:DeblurGAN:2018}. 

Theoretically, the generator in a GAN will not learn to create new content, but it will just try to make its output look like the real data. Therefore, to produce creative works of art, the Creative Adversarial Network (CAN) has been proposed by \citet{Elgammal:CAN:2017}. This works by including an additional signal in the generator to prevent it from generating content that is too similar to existing examples. Similar to traditional CNNs, a perceptual loss based on VGG16 \citep{Johnson:Perceptual:2016} has become common in applications where new images are generated that have the same semantics as the input \citep{Ledig:Photo:2017,Antic:DeOldify:2020}. 

Most GAN-based methods are currently limited to the generation of relatively small square images, e.g., 256$\times$256 pixels \citep{Zhang:StackGAN:2017}. The best resolution created up to the time of this review is 1024$\times$1024-pixels, achieved by NVIDIA research. The team introduced the progressive growing of GANs \citep{Karras:Progressive:2018} and showed that their method can generate near-realistic 1024$\times$1024-pixel portrait images (trained for 14 days). However the problem of obvious artefacts at transition areas between foreground and background persists.

Another form of deep generative model is the Variational Autoencoder (VAE). A VAE is an autoencoder, where the encoding distribution is regularised to ensure the latent space has good properties to support the generative process. Then the decoder samples from this distribution to generate new data. Comparing VAEs to GANs, VAEs are more stable during training, {while} GANs are better at producing realistic images. Recently Deepmind (Google) has included vector quantization (VQ) within a VAE to learn a discrete latent representation  \citep{Razavi:VQVAE:2019}. Its performance for image generation are competitive with their BigGAN \citep{Brock:Large:2019} but with greater capacity for generating a diverse range of images. There have also been many attempts to merge GANs and VAEs so that the end-to-end network benefits from both good samples and good representation, for example using a VAE as the generator for a GAN \citep{Wan:Crossing:2017, Bhattacharyya:Best:2019}. However, the results of this have not yet demonstrated significant improvement in terms of overall performance \citep{Rosca:Distribution:2019}, remaining an ongoing research topic.

A review of recent state-of-the-art GAN models and applications can be found in \citep{Foster:Generative:2019}.

\subsubsection{Recurrent Neural Networks (RNNs)}

Recurrent neural networks (RNNs) have been widely employed to perform sequential recognition; they offer benefits in this respect by incorporating at least one feedback connection. The most commonly used type of RNN is the Long Short-Term Memory (LSTM) network \citep{Hochreiter:Long:1997}, as this solves problems associated with  vanishing gradients, observed in traditional RNNs. It does this by memorizing sufficient context information in time series data via its memory cell. Deep RNNs use their internal state to process variable length sequences of inputs, combining across multiple levels of representation. This makes them amenable to tasks such as speech recognition \citep{Graves:speech:2013},  handwriting recognition \citep{Doetsch:Fast:2014}, and music generation \citep{Briot:Deep:2020}. 
RNNs are also employed in image and video processing applications, where recurrency is applied to convolutional encoders for tasks such as  drawing sketches \citep{Ha:Neural:2017} and deblurring videos \citep{Zhang:Dynamic:2018}. VINet \citep{Kim:Deep:2019} employs an encoder-decoder model using an RNN to estimate optical flow, processing multiple input frames concatenated with the previous inpainting results. An example network using an RNN is illustrated in Fig. \ref{fig:GANdiagrams} (b).

{
CNNs extract spatial features from its input images using convolutional filters and RNNs extract sequential features in time-series data using memory cells. In extension, 3D CNN, CNN-LSTM and ConvLSTM have been designed to extract spatial-temporal features from video sequences. The 3D activation maps produced in 3D CNNs are able to analyze  temporal or volumetric context which are important in applications such as medical imaging \citep{Selvikvag:overview:2019} and action recognition \citep{Ji:3D:2013}. The CNN-LSTM simply concatenates a CNN and an LSTM (the 1D output of the CNN is the input to the LSTM) to process time-series data. In contrast, ConvLSTM is another LSTM variant, where the internal matrix multiplications are replaced with convolution operations at each gate of the LSTM cell so that the LSTM input can be in the form of multi-dimensional data \citep{Shi:convLSTM:2015}. 
}

\subsubsection{Deep Reinforcement Learning (DRL)}
Reinforcement learning (RL) is an ML algorithm trained to make a sequence of decisions.  Deep reinforcement learning (DRL) combines ANNs with an RL architecture that enables RL agents to learn the best actions in a virtual environment to achieve their goals. The RL agents are comprised of a policy that performs a mapping from an input state to an output action and an algorithm responsible for updating this policy. This is done through leveraging a system of rewards and punishments to acquire useful behaviour -- effectively a trial-and-error process. The framework trains using a simulation model, so it does not require a predefined training dataset, either labeled or unlabeled. 

However, pure RL requires an excessive number of trials to learn fully, something that may be impractical in many (especially real-time) applications if training from scratch \citep{Hessel:Rainbow:2018}. AlphaGo, a computer {program} developed by DeepMind Technologies that can beat a human professional Go player, employs RL on top of a pre-trained model to improve its play strategy to beat a particular player\footnote{https://ai.googleblog.com/2016/01/alphago-mastering-ancient-game-of-go.html}. RL could be useful in creative applications, where there may not be a predefined way to perform a given task, but where there are rules that the  model has to follow to perform its duties correctly.  Current applications involve end-to-end RL combined with CNNs, including gaming \citep{Mnih:Playing:2013}, and RLs with GANs in optimal painting stroke in stroke-based rendering \citep{Huang:Learning:2019}. Recently RL methods have been developed using a graph neural network (GNN) to play Diplomacy, a highly complex 7-player (large scale) board game \citep{Anthony:Learning:2020}.

Temporal difference (TD) learning \citep{Gregor:Temporal:2018,Chen:temporal:2018, Nguyen:Deep:2020} has recently been introduced as a model-free reinforcement learning method that learns how to predict a quantity that depends on future values of a given signal. That is, the model learns from an environment through episodes with no prior knowledge of the environment. This may well have application in the creative sector for storytelling, caption-from-image generation and gaming.

\section{AI for the Creative Industries}
\label{sec:exiting}

AI has increasingly (and often mistakenly) been associated with human creativity and artistic practice. As it has demonstrated abilities to `see', `hear', `speak', `move', and `write', it has been applied in domains and applications including: audio, image and video analysis, gaming, journalism, script writing, filmmaking,  social media analysis and marketing. One of the earliest AI technologies, available for more than two decades, is Autotune, which automatically fixes vocal intonation errors \citep{Hildebrand:Pitch}. An early attempt to exploit AI for creating art occurred in 2016, when a three-dimensional (3D) printed painting, the Next Rembrandt\footnote{https://www.nextrembrandt.com/}, was produced solely based on training data from Rembrandt’s portfolio. It was created using deep learning algorithms and facial recognition techniques. 

Creativity is defined in the Cambridge Dictionary as `the ability to produce original and unusual ideas, or to make something new or imaginative'. Creative tasks generally require some degree of original thinking, extensive experience and an understanding of the audience, {while} production tasks are, in general, more repetitive or predictable, making them more amenable to being performed by machines. To date, AI technologies have produced mixed results when used for generating original creative works. For example, GumGum\footnote{https://gumgum.com/artificial-creativity} creates a new piece of art following the input of a brief idea from the user. The model is trained by recording the preferred tools and processes that the artist uses to create a painting. A Turing test revealed that it is difficult to distinguish these AI generated products from those painted by humans.  AI methods often produce unusual results when employed to create new narratives for books or movie scripts. Botnik\footnote{https://botnik.org/} employs an AI algorithm to automatically remix texts of existing books to create a new chapter. In one experiment, the team fed the seven Harry Potter novels through their predictive text algorithm, and the `bot' created rather strange but amusing sentences, such as ``\textit{Ron was standing there and doing a kind of frenzied tap dance. He saw Harry and immediately began to eat Hermione’s family}" \citet{Sautoy:Creativity:2019}. However, when AI is used to create less structured content (e.g., some forms of `musical' experience), it can demonstrate pleasurable difference \citep{Briot:Deep:2020}.

In the production domain, Twitter has applied automatic cropping to create image thumbnails that show the most salient part of an image \citep{Theis:Faster:2018}. The BBC has created a proof-of-concept system for automated coverage of live events. In this work, the AI-based system performs shot framing (wide, mid and close-up shots), sequencing, and shot selection automatically \citep{Wright:AI:2020}. However, the initial results show that the algorithm needs some improvement if it is to replace human operators. Nippon Hoso Kyokai (NHK, Japan's Broadcasting Corporation), has developed a new AI-driven broadcasting technology called ``Smart Production". This approach extracts events and incidents from diverse sources such as social media feeds (e.g., Twitter), local government data and interviews, and integrates these into a human-friendly accessible format   \citep{Kaneko:AI:2020}.

In this review, we divide creative applications into five major categories: content creation, information analysis, content enhancement and post production workflows, information extraction and enhancement, and data compression. However, it should be noted that many applications exploit several categories in combination.  
For instance, post-production tools (discussed in Section \ref{ssec:contentenhance} and \ref{ssec:infoextract}) frequently combine  information extraction and content enhancement techniques.  These combinations can together be used to create new experiences, enhance existing material  or to re-purpose archives (e.g., `Venice Through a VR Lens, 1898' directed by BDH Immersive and Academy 7 Production\footnote{https://www.bdh.net/immersive/venice-1898-through-the-lens.}). These workflows may employ AI-enabled super-resolution, colorization, 3D reconstruction and frame rate interpolation methods.
{Gaming is another important example that has been key for the development of AI. It could be considered as an `all-in-one' AI platform, since it combines rendering, prediction and learning.}

We {categorize} the applications and the corresponding AI-based solutions as shown  in Table \ref{tab:gather}. For those interested, a more detailed overview of contemporary Deep Learning systems is provided in Section \ref{sec:methods}. 

\begin{landscape}
\begin{table}
\caption{Creative applications and corresponding AI-based methods}
\tiny
 \hskip-5.0cm\begin{tabular}{p{1cm}p{1.4cm}|p{7cm}p{4cm}p{4cm}p{4cm}}
 \\
 \toprule
\multicolumn{2}{c}{\multirow{2}{*}{Application}} & \multicolumn{4}{c}{Technology} \\ \cline{3-6}
& & CNN & GAN & RNN  & other\\
\hline
{\bf Creation} & Content generation for text, audio, video and game & \citet{Zhang:colorful:2016,Gatys:Neural:2016,Hessel:Real:2019,Zhang:design:2020,Quesnel:deep:2018} & \citet{Radford:Unsupervised:2016, Isola:Image:2017, Jin:Towards:2017, Zhu:Unpaired:2017, Yi:DualGAN:2017, Wang:vid2vid:2018, Subramanian:towards:2018, Li:Generative:2018, Karras:Progressive:2018, Song:Dual:2018,  Zakharov:Few:2019, Li:Automatic:2019, Brock:Large:2019, Kim:GameGAN:2020, donahue:wavegan:2019,Engel:GANSynth:2019, He:AttGAN:2019,Song:Dual:2018,He:AttGAN:2019} & \citet{Subramanian:towards:2018, Li:Automatic:2019,Venugopalan:Translating:2015, Sturm:Music:2016,Ha:Neural:2017,Mao:DeepJ:2018, Bahdanau:Neural:2018,Kim:GameGAN:2020}  & TM \citep{Dorr:Mapping:2016}, RL \citep{Wang:Interactive:2017,Gregor:Temporal:2018,Chen:temporal:2018,Nguyen:Deep:2020,Chen:SoundSpaces:2020}, BERT \citep{Devlin:BERT:2019}, VAE \citep{Razavi:VQVAE:2019}, NEAT \citep{Stanley:HyperNEAT:2009}, Graph-based  \citep{Chaplot:TopologicalSLAM:2020}\\
& Animation & \citet{Starke:Local:2020, Holden:Learning:2015, Starke:Neural:2019, Nalbach:DeepShading:2017, Tesfaldet:Two:2018} &  \citet{Nagano:paGAN:2018, Wei:VR:2019} & \citet{Lee:Interactive:2018,Siyao:deepanimation:2021}  & VAE  \citep{Wei:VR:2019}\\
 & AR/VR & \citet{Panphattarasap:automated:2018,Anantrasirichai:fixation:2016} & \\
& Deepfakes & & \citet{Chan:Everybody:2019} & \citet{Suwajanakorn:obame:2017} & VAE \citep{Kietzmann:deepfakes:2020} \\
& Content and captions & \citet{Pu:Variational:2016,Chen:Order:2018, Chen:multi:2019} & \citet{Mansimov:Generating:2016,Zhang:StackGAN:2017,Li:Object:2019} & \citet{Xia:recurrent:2005,Xu:Learnging:2017,Chen:Order:2018} & VAE \citep{Pu:Variational:2016}\\
\hline
{\bf Information Analysis} & Text categorization & \citet{Johnson:Effective:2015} & \citet{ Li:Generative:2018} & \citet{Chen:Ensemble:2017,Gunasekara:review:2018,Trusca:hybrid:2020} & SOM \citep{Pawar:Comparative:2012}, ELM \citep{Mohammad:Regularizing:2021}\\
& Ads/film analysis & \citet{Young:recent:2018} & \citet{Young:recent:2018} & \citet{Young:recent:2018} & GP \citep{Lacerda:Learning:2006} \\
 & Content retrieval & \citet{Wu:Deep:2015, Amato:searching:2017,Wan:DeepRetrieval:2014, Gordo:DeepRetrieval:2016} & \citet{Song:Binary:2018} & \citet{Jabeen:Video:2018} & PM \citep{Jeon:Automatic:2003} \\
 & Fake detection & \citet{Guera:deepfake:2018, Li:Exposing:2019} & & \citet{Guera:deepfake:2018} & Blockchain \citep{Hasan:combating:2019} \\
 & Recommendation & \citet{Deldjoo:audio:2018} & & \citet{rush:nueral:2015,See:get:2017,Yi:Personalized:2020,Li:Towards:2019} & Deep belief net \citep{Batmaz:review:2019}, regularization \citep{Chen:Resource:2014} \\
 \hline
{\bf Content} & Contrast  & \citet{Lore:LLNet:2017} & \citet{Kuang:single:2019, Jiang:EnlightenGAN:2021,Anantrasirichai:Contextual:2021} &  & Histogram \citep{Pizer:adaptive:1987} \\
{\bf  Enhancement and Post} & Colorization & \citet{Cheng:Colorization:2015, Ronneberger:Unet:2015, Limmer:Infrared:2016, Xu:Stylization:2020}  & \citet{Zhang:seftattention:2019, Antic:DeOldify:2020, Suarez:Infrared:2017,Kuang:Thermal:2020, Anantrasirichai:Contextual:2021} & & \\
{\bf  Production} & Super-resolution & \citet{Dong:Learning:2014, Huang:Bidirectional:2015, Shi:Real:2016, Kappeler:video:2016, Kim:Accurate:2016,Tai:SR:2017, Caballero:realtime:2017,Dai:Deformable:2017, Sajjadi:EnhanceNet:2017, Zhang:image:2018, Liu:Learning:2018, Wang:EDVR:2019, Dai:Second:2019, Wang:SR:2019, Haris:Recurrent:2019} & \citet{Ledig:Photo:2017, Wang:Fully:2018} & \citet{Huang:Bidirectional:2015, Haris:Recurrent:2019} &\\
 & {Deblurring} & \citet{Hradis:text:2015, Schuler:Learning:2016, Nah:Deep:2017, Su:Deep:2017, Kim:Online:2017, Tao:scale:2018, Zhang:Dynamic:2018, Gao:dynamic:2019, Zhou:Spatio:2019} &\citet{Kupyn:DeblurGAN:2018} & \citet{Tao:scale:2018},  & Statistic model \citep{Biemond:Iterative:1990, Kim:Online:2017,Zhang:Dynamic:2018,Nah:Recurrent:2019}, BD \citep{Jia:single:2007, Krishnan:Blind:2011} \\
 & Denoising & \citet{Zhang:dncnn:2017, Xiaoming:Learning:2018, Liu:Multi:2018, Zhang:FFDNet:2018, Claus:ViDeNN:2019, Davy:nonlocal:2019, Xue:Video:2019, Zhao:simple:2019, Chen:image:2018, Lehtinen:noise2noise:2018, Krull:Noise2Void:2019,Chen:Learning:2018, Lempitsky:Deep:2018, Brooks:Unprocessing:2019, Xu:Deformable:2019} & \citet{Chen:image:2018, Yang:Low:2018} & \citet{Maas:Recurrent:2012, Zhang:Deep:2018,Anantrasirichai:Contextual:2021} & Filtering \citep{Yahya:video:2016, Malm:adaptive:2007, Maggioni:BM4D:2012, Zuo:video:2013, Buades:CFA:2019} \\
 & Dehazing & \citet{Cai:DehazeNet:2016, Li:Underwater:2016, Li:AOD:2017, Yang:Proximal:2018, Hu:Underwater:2018} & \citet{Engin:cycle:2018, Tang:single:2019} \\
 & {Turbulence removal} & { \citet{Gao:Atmospheric:2019, Nieuwenhuizen:deep:2019}} & { \citet{Chak:Subsampled:2018}} & &  Fusion \citep{Anantrasirichai:Atmospheric:2013},  \\
  & & & &  & BD \citep{Zhu:Removing:2013, Xie:Removing:2016} \\
 & Inpainting & \citet{Xie:Image:2012,Kim:Deep:2019, Hong:Deep:2019} & \citet{Yu:Generative:2018, Yu:Freeform:2019, Chang:Free:2019} &  \citet{Kim:Deep:2019} & Sparse coding \citep{Xie:Image:2012} \\
 & VFX & \citet{Hu:Avatar:2017,Torrejon:rotoscope:2020} & & & Filtering \citep{Barber:camera:2016}\\
 \hline
{\bf Information Extraction} & Segmentation & \citet{Long:fully:2015, Noh:Learning:2015, Ronneberger:Unet:2015, Kirillov:PointRend:2020, Taghanaki:semantic:2021, Qi:PointNet:2017}  &  \citet{Isola:Image:2017} & & \\
 & Recognition & \citet{Ji:3D:2013,Wang:TSN:2016,Ren:FasterRCNN:2017,He:Mask:2017,Redmon:YOLO:2016, Redmon:YOLOv3:2018, Yang:NAS:2020, Shillingford:Large:2019, Kazakos:EPIC:2019,Cai:Flattenet:2019, Zhen:Learning:2019, Peng:MegDet:2018, Chen:mmdet:2019, Sun:FishNet:2018,Kim:Learning:2020,Adithya:deep:2020, Jalal:American:2018}  &  & \citet{Shillingford:Large:2019} & \\
  & Tracking & \citet{Li:High:2018, Wang:Fast:2019, Bochkovskiy:YOLOv4:2020, Ciaparrone:Deep:2020, Borysenko:ODESA:2020} & & \citet{Fang:Track:2016,Milan:Online:2017,Gordon:Re3:2018} \\
  & SOD & \citet{Hou:Deeply:2019, Fan:shifting:2019, Fan:taking:2020} & \citet{Mejjati:Parametric:2020, Jiang:cmSalGAN:2020, Wang:SaliencyGAN:2020} & \citet{Fan:shifting:2019} & graph-cut \citep{Cheng:RepFinder:2010}, multi-scale features \citep{Gupta:visual:2013} \\
  & Fusion & \citet{Liu:multi:2017, Prabhakar:DeepFuse:2017} & \citet{Ma:FusionGAN:2019, Wu:GPGAN:2019} & & Filtering \citep{Ma:Infrared:2019, Li:Image:2013, Anantrasirichai:Image:2020} \\
& 3D Reconstruction  & \citet{Newell:hourglass:2016,Chang:pyramid:2018,Zhang:GANet:2019,Anantrasirichai:Fast:2020, Bulat:How:2017, Jackson:3Dface:2017, Tewari:High:2020,Kanazawa:End:2018, Cheng:Learning:2019, Xie:Pix2Vox:2019,Flynn:DeepView:2019, Mescheder:Occupancy:2019, Gao:25Dsound:2019, Gkioxari:mesh:2019,Shi:learning:2020,Morgado:self:2018, Vasudevan:Semantic:2020} & \citet{Yang:3D:2017,Jiang:GAL:2018,Tian:CRGAN:2018,Shimada:IsMo:2019,Wu:MarrNet:2017} & & VAE \citep{Soltani:Synthesizing:2017} \\
\hline
{\bf Compression} & & \citet{Jiang:compression:2018, Han:Deep:2019, Zhang:Enhancing:2020,Stankiewicz:video:2019,KuoTCSVT2009,Ma:Perceptually:2019, Zhang:ViSTRA2:2019, Schiopu:CNN:2019,Zhao:Enhanced:2019, Liu:CNN:2018,Li:Fully:2018,Xue:Attention:2019, Lu:end:2020} & \citet{Ma:Perceptually:2019, Ma:GAN:2020}  & \citet{Goyal:DeepZip:2019} & VAE \citep{Han:Deep:2019}\\
\bottomrule
\\
\multicolumn{6}{l}{CNN: Convolutional Neural Network, GAN: Generative Adversarial Network} \\
\multicolumn{6}{l}{RNN: Recurrent Neural Network, RL: Reinforcement Learning, PM: Probabilistic model }\\
\multicolumn{6}{l}{BERT: Bidirectional Encoder Representations from Transformers, TM: Text mining} \\
\multicolumn{6}{l}{VAE: Variational Autoencoders, AR: Augmented Reality, VR: Virtual Reality}\\
\multicolumn{6}{l}{GP: Genetic Programming, BD: Blind Deconvolution, VFX: Visual Effects}, SOM: Self-Organizing Map\\
\multicolumn{6}{l}{NEAT: NeuroEvolution of Augmenting Topologies, ELM: Extreme Learning Machine} \\
 \end{tabular}
\label{tab:gather}
\end{table}
 \end{landscape}

\subsection{Content Creation}
\label{ssec:contentcreation}

   Content creation is a fundamental activity of artists and designers. This section discusses how AI technologies have been employed both to support the creative process and as a creator in their own right. 

\subsubsection{Script and Movie Generation}

\begin{figure*}[t!]
	\centering
	\vspace{5mm}
      		 \includegraphics[width=\textwidth]{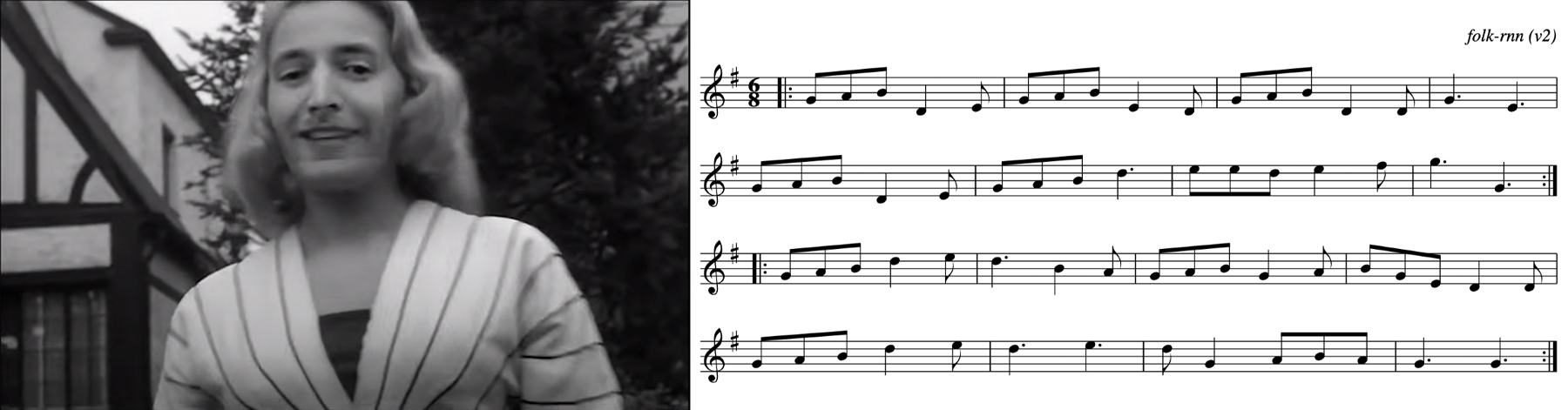} 
      		 \begin{tabular}{x{5cm}x{6.5cm}}
			(a) & (b)
\end{tabular} 
					\caption{\small (a) A screenshot from `Zone Out', where the face of the woman was replaced with a man's mouth$^{\ref{fn:zoneout}}$. (b) Music transcription generated by AI algorithm$^{\ref{Sturnchannel}}$}.
    \label{fig:scriptget}
\end{figure*}

The narrative or story underpins all forms of  creativity across art, fiction,  journalism, gaming, and other forms of entertainment. AI has been used both to create  stories and to optimize the use of supporting data, for example {organizing} and searching through huge archives for documentaries. The script of a fictional short film, Sunspring (2016)\footnote{https://www.imdb.com/title/tt5794766/}, was entirely written by an AI machine, known as Benjamin, created by New York University. The model, based on a recurrent neural network (RNN) architecture, was trained using science fiction screenplays as input, and the script was generated with random seeds from a sci-fi filmmaking contest. Sunspring has some unnatural story lines. In the sequel, It's No Game (2017),  Benjamin was then used only in selected areas and in collaboration with humans, producing a more fluid and natural plot. This reinforces the notion  that the current AI technology can work more efficiently in conjunction with humans rather than being left to its own devices. In 2016, IBM Watson, an AI-based computer system, composed the 6-min movie trailer of a horror film, called Morgan\footnote{https://www.ibm.com/blogs/think/2016/08/cognitive-movie-trailer/}. The model was trained with more than 100 trailers of horror films enabling it to learn the normative structure and pattern. Later in 2018, Benjamin was used to generate a new film `Zone Out' (produced within 48 hours). The project also experimented further by using face-swapping, based on a GAN and voice-generating technologies. This film was entirely directed by AI, but includes many artefacts and unnatural scenes as shown in Fig. \ref{fig:scriptget} (a)\footnote{https://www.youtube.com/watch?v=vUgUeFu2Dcw \label{fn:zoneout}}. Recently, ScriptBook\footnote{https://www.scriptbook.io}  introduced a story-awareness concept for AI-based storytelling. The generative models focus on three aspects: awareness of characters and their traits, awareness of a script’s style and theme, and awareness of a script’s structure, so the resulting script is more natural.

In gaming, AI has been used to support design, decision-making and interactivity \citep{Justesen:deep:2020}. Interactive narrative, where users create a storyline through actions, has been developed using AI methods over the past decade \citep{Riedl:Interactive:2012}. For example, MADE (Massive Artificial Drama Engine for non-player characters) generates procedural content in games \citep{Hector:MADE:2014}, and deep reinforcement learning has been employed for personalization \citep{Wang:Interactive:2017}. AI Dungeon\footnote{https://aidungeon.io/} is a web-based game that is capable of generating a storyline in real time, interacting with player input. The underlying algorithm requires more than 10,000 label contributions for training to ensure that the model produces smooth interaction with the players. 
Procedural generation has been used to automatically randomize content so that a game does not present content in the same order every time \citep{Short:Procedural:2017}. Modern games often integrate 3D visualization, augmented reality (AR) and virtual reality (VR) techniques, with the aim of making play  more realistic and immersive. Examples include Vid2Vid \citep{Wang:vid2vid:2018} which uses a deep neural network, trained on real videos of cityscapes, to generate a synthetic 3D gaming environment. Recently, NVIDIA Research has used a generative model (GameGAN by \citet{Kim:GameGAN:2020}), trained on 50,000 PAC-MAN episodes, to create new content, which can be used by game developers to automatically generate layouts for new game levels in the future. 

\subsubsection{Journalism and text generation}

Natural language processing (NLP) refers to the broad class of computational techniques for incorporating speech and text. It analyzes natural language data and trains machines to perceive and to generate human language directly. NLP algorithms frequently involve speech recognition (Section \ref{sssec:recog}), natural language understanding (e.g., BERT by Google AI \citep{Devlin:BERT:2019}), and natural language generation \citep{leppanen:data:2017}.  
Automated journalism, also known as robot journalism, describes automated tools that can generate news articles from structured data. The process scans large amounts of assorted data, orders key points, and inserts details such as names, places, statistics, and some figures \citep{Cohen:From:2015}. This can be achieved through NLP and text mining techniques \citep{Dorr:Mapping:2016}. 

AI can help to break down barriers between different languages with machine translation \citep{Bahdanau:Neural:2018}.  A conditioned GAN with an RNN architecture has been proposed for language translation by \citet{Subramanian:towards:2018}. It was used for the difficult task of generating English sentences from Chinese poems; it creates understandable text but sometimes with grammatical errors. CNN and RNN architectures are employed to translate video into natural language sentences  \citep{Venugopalan:Translating:2015}. AI can also be used to rewrite one article to suit several different channels or audience tastes\footnote{https://www.niemanlab.org/2016/10/the-ap-wants-to-use-machine-learning-to-automate-turning-print-stories-into-broadcast-ones/}. A survey of recent deep learning methods for text generation by \citet{Iqbal:survey:2020} concludes that text {generated from images could be most} amenable to  GAN processing while topic-to-text translation is likely to be dominated by variational autoencoders (VAE). 

Automated journalism is now quite widely used. For example, BBC reported on the UK general election in 2019 using such tools\footnote{https://www.bbc.com/news/technology-50779761}. Forbes uses an AI-based content management system, called Bertie, to assist in providing reporters with the first drafts and templates for news stories\footnote{https://www.forbes.com/sites/nicolemartin1/2019/02/08/did-a-robot-write-this-how-ai-is-impacting-journalism/\#5292ab617795}. The Washington Post also has a robot reporting program called Heliograf\footnote{https://www.washingtonpost.com/pr/wp/2016/10/19/the-washington-post-uses-artificial-intelligence-to-cover-nearly-500-races-on-election-day/}. Microsoft has announced in 2020 that they use automated systems to select news stories to appear on MSN website\footnote{https://www.bbc.com/news/world-us-canada-52860247}.
This application of AI demonstrates that current AI technology can be effective in supporting human journalists in constrained cases,  increasing production efficiency.

\subsubsection{Music Generation}

There are many different areas where sound design is used in professional practice, including television, film, music production, sound art, video games and theatre. Applications of AI in this domain include searching through large databases to find the most appropriate match  for such applications (see Section \ref{sssec:retrieval}), and assisting sound design. Currently, several AI assisted music composition systems support music creation.  The process generally involves using ML algorithms to analyze data to find musical patterns, e.g., chords, tempo, and length from  various instruments, synthesizers and drums. The system then suggests new composed melodies that may inspire the artist.  Example software includes Flow Machines by Sony\footnote{http://www.flow-machines.com/}, Jukebox by OpenAI\footnote{https://openai.com/blog/jukebox/} and NSynth by Google AI\footnote{https://magenta.tensorflow.org/nsynth}.  In 2016, Flow Machines launched a song in the style of The Beatles, and in 2018 the team released the first AI album, `Hello World', composed by an artist, SKYGGE (Benoit Carr\'{e}), using an AI-based tool\footnote{https://www.helloworldalbum.net/}. {Coconet uses a CNN to infill  missing pieces of music\footnote{https://magenta.tensorflow.org/coconet}}.
Modelling music creativity is often achieved using Long Short-Term Memory (LSTM), a special type of RNN architecture \citep{Sturm:Music:2016} (an example of the  output of this model is shown in Fig. \ref{fig:scriptget} (b)\footnote{https://folkrnn.org/} and the reader can experience AI-based music at Ars Electronica Voyages Channel\footnote{https://ars.electronica.art/keplersgardens/en/folk-algorithms/ \label{Sturnchannel}}). The model takes a transcribed musical idea and transforms it in meaningful ways. For example, DeepJ composes music conditioned on a specific mixture of composer styles using a Biaxial LSTM architecture \citep{Mao:DeepJ:2018}. More recently,  generative models have been configured based on an LSTM neural network to generate music \citep{Li:Automatic:2019}.   

Alongside these methods of musical notation based audio synthesis, there also exists a range of direct waveform synthesis techniques that learn and/or act directly on the waveform of the audio itself (for example  \citep{donahue:wavegan:2019,Engel:GANSynth:2019}.  A more detailed overview of Deep Learning techniques for music generation can be found in \citep{Briot:Deep:2020}. 

\begin{figure*}[t!]
	\centering
	\vspace{5mm}
      		 \includegraphics[width=\textwidth]{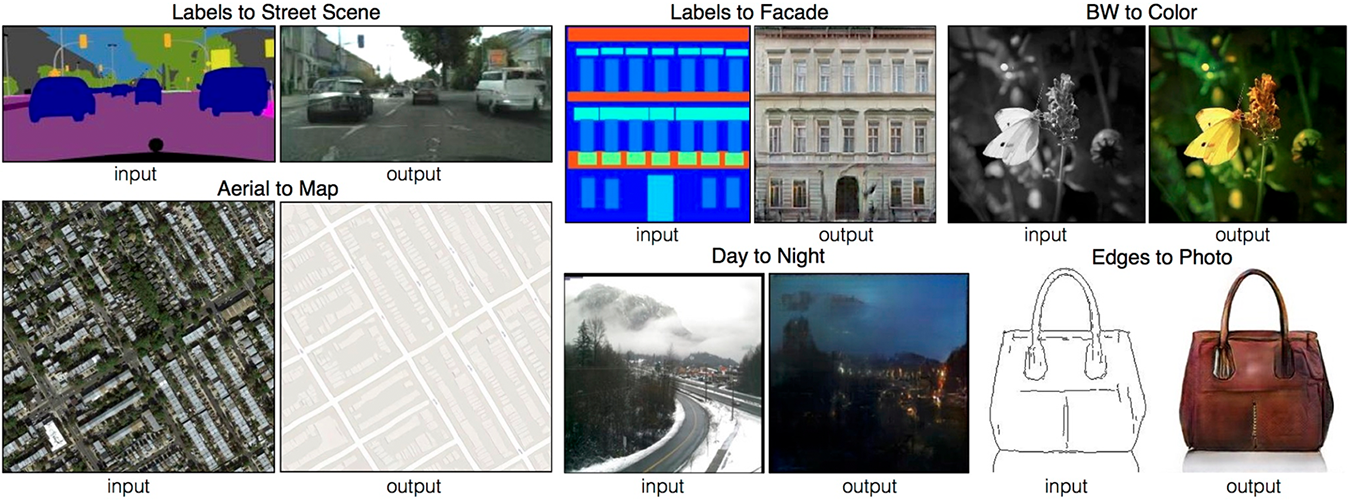}  
					\caption{\small Example applications of pix2pix framework \citep{Isola:Image:2017}}
    \label{fig:pix2pix}
\end{figure*} 

\subsubsection{Image Generation}
\label{sssec:imggen}
AI can be used to create new digital imagery or art-forms automatically, based on selected training datasets, e.g., new examples of bedrooms \citep{Radford:Unsupervised:2016}, cartoon characters \citep{Jin:Towards:2017}, celebrity headshots \citep{Karras:Progressive:2018}. Some applications produce a new image conditioned to the input image, referred to as image-to-image translation, or `\textit{style transfer}'. It is called translation or transfer, because the image output has a different appearance to the input but with similar  semantic content. That is, the algorithms learn the mapping between an input image and an output image.
For example, grayscale tones can be converted into natural colors \citep{Zhang:colorful:2016}, using eight simple convolution layers to capture {localize}d semantic meaning and to generate $a$ and $b$ color channels of the CIELAB color space. This involves mapping class probabilities to point estimates in $ab$ space. DeepArt \citep{Gatys:Neural:2016}  transforms the input image into the style of the selected artist by combining feature maps from different convolutional layers. A stroke-based drawing method trains machines to draw and generalise abstract concepts in a manner similar to humans using RNNs  \citep{Ha:Neural:2017}. 

A Berkeley AI Research team has successfully used GANs to convert between two image types \citep{Isola:Image:2017}, e.g., from a Google map to an aerial photo, a segmentation map to a real scene, or a sketch to a colored object (Fig. \ref{fig:pix2pix}). They have published their pix2pix codebase\footnote{https://phillipi.github.io/pix2pix/} and invited the online community to experiment with it in different application domains, including depth map to street view, background removal and pose transfer. For example pix2pix has been used\footnote{https://ai-art.tokyo/en/} to create a Renaissance portrait from a real portrait photo. Following pix2pix, a large number of research works have improved the performance of style transfer.  Cycle-Consistent Adversarial Networks (CycleGAN) \citep{Zhu:Unpaired:2017} and DualGAN \citep{Yi:DualGAN:2017} have been proposed for unsupervised learning. Both algorithms are based on similar concepts -- the images of both groups are translated twice (e.g., from group A to group B, then translated back to the original group A) and the loss function compares the input image and its reconstruction, computing what is referred to as cycle-consistency loss. Samsung AI has shown, using GANs, that it is possible to turn a portrait image, such as the Mona Lisa, into a video where the portrait's face speaks in the style of a guide  \citep{Zakharov:Few:2019}. Conditional GANs can be trained to transform a human face into one of a different age \citep{Song:Dual:2018}, and to change facial attributes, such as the presence of a beard, skin condition, hair style and color \citep{He:AttGAN:2019}.

Several creative tools have employed ML-AI methods to create new unique artworks. For example, Picbreeder\footnote{http://picbreeder.org/} and EndlessForms\footnote{http://endlessforms.com/} employ Hypercube-based NeuroEvolution of Augmenting Topologies \citep{Stanley:HyperNEAT:2009} as a generative encoder that exploits geometric regularities.  Artbreeder\footnote{https://www.artbreeder.com/} and GANVAS Studio\footnote{https://ganvas.studio/} employ BigGAN \citep{Brock:Large:2019} to generate high-resolution class-conditional images and also to mix two images together to create new interesting work. 

\begin{figure*}[t!]
	\centering
	\vspace{5mm}
      		 \includegraphics[width=\textwidth]{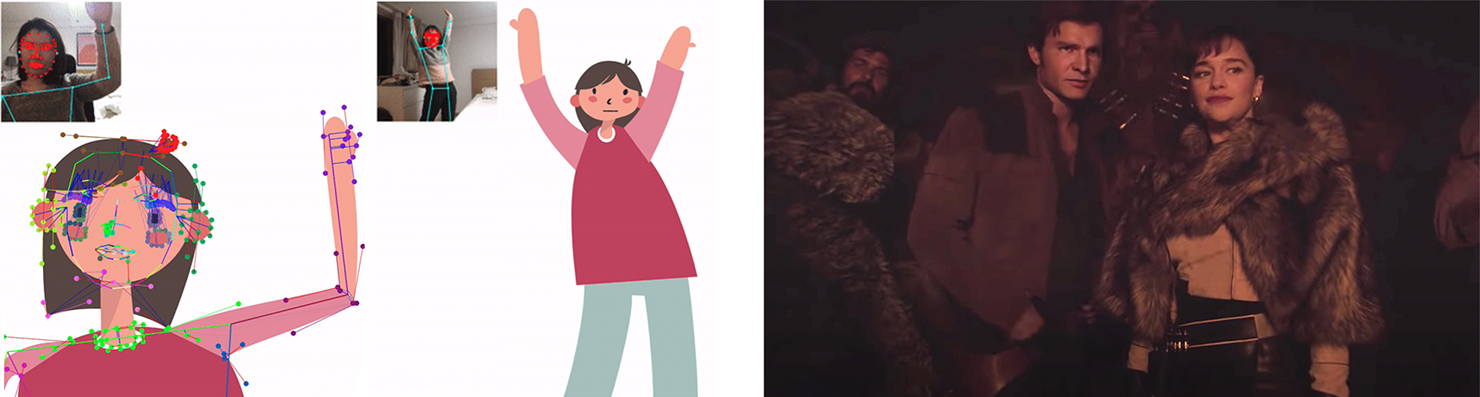}  
      		 \begin{tabular}{x{5.8cm}x{6cm}}
			(a) & (b)
\end{tabular} 
					\caption{\small (a) Real-time pose animator$^{\ref{fn:poseanimator}}$. (b) Deepfake applied to replaces Alden Ehrenreich with young Harrison Ford  in Solo: A Star Wars Story by derpfakes$^{\ref{fn:derpfakes}}$}
    \label{fig:poseanimation}
\end{figure*}

\subsubsection{Animation}

Animation is the process of using drawings and models to create moving images. Traditionally this was done by hand-drawing each frame in the sequence and rendering these at an appropriate rate to give the appearance of continuous motion. {In recent years,} AI methods have been employed to automate the animation process making it easier, faster and more realistic than in the past. A single animation project can involve several shot types, ranging from simple camera pans on a static scene, to more challenging dynamic movements of multiple interacting characters (e.g basketball players  \citep{Starke:Local:2020}).   ML-based AI is particularly well suited to learning models of motion from captured real motion sequences. These motion characteristics can be learnt using deep learning-based approaches, such as  autoencoders \citep{Holden:Learning:2015},  LSTMs \citep{Lee:Interactive:2018}, and motion prediction networks \citep{Starke:Neural:2019}. Then, the inference applies these characteristics from the trained model to animate characters and dynamic movements.
In simple animation, the motion can be estimated using a single low-cost camera. For example, Google research has created software for pose animation that turns a human pose into a cartoon animation in real time\footnote{https://github.com/yemount/pose-animator/ \label{fn:poseanimator}}. This is based on PoseNet (estimating pose position\footnote{https://www.tensorflow.org/lite/models/pose\_estimation/overview}) and FaceMesh (capturing face movement \citep{Hessel:Real:2019}) as shown in Fig. \ref{fig:poseanimation} (a). Adobe has also created Character Animator software\footnote{https://www.adobe.com/products/character-animator.html} offering lip synchronisation, eye tracking and gesture control through webcam and microphone inputs in real-time. This has been adopted by Hollywood studios and other online content creators. 

AI has also been employed for rendering objects and scenes. This includes the synthesis of 3D views from motion capture or from monocular cameras (see Section \ref{sssec:3Dreconstruct}), shading \citep{Nalbach:DeepShading:2017}  and dynamic texture synthesis \citep{Tesfaldet:Two:2018}. Creating realistic lighting in animation and visual effects has also benefited by combining traditional geometrical computer vision with enhanced ML approaches and multiple depth sensors  \citep{Guo:Relightables:2019}. Animation is not only important within the film industry; it also plays an important role in the games industry, responsible for the portrayal of movement and behaviour. Animating characters, including their faces and postures, is a key component in a game engine. AI-based technologies have enabled digital characters and audiences to co-exist and interact\footnote{https://cubicmotion.com/persona/}. Avatar creation has also been employed to enhance virtual assistants\footnote{https://www.pinscreen.com/}, e.g., using proprietary photoreal AI face synthesis technology \citep{Nagano:paGAN:2018}. Facebook Reality Labs have employed ML-AI techniques to animate realistic digital humans, called Codec Avatars,  in real time using GAN-based style transfer and using a VAE to extract avatar parameters \citep{Wei:VR:2019}. AI is also employed to up-sample frame rate in animation\cite{Siyao:deepanimation:2021}.

\subsubsection{Augmented, Virtual and Mixed Reality (VR, AR, MR)}
 
AR and VR use computer technologies to create a fully simulated  environment or one that is real but augmented with virtual entities. AR expands the physical world with digital layers via mobile phones,  tablets or head mounted displays, while VR takes the user into immersive experiences via a headset with a 3D display that isolates the viewer (at least in an audio-visual sense) from the physical world {\citep{Milgram:Augmented:1995}}. 

Significant predictions have been made about the growth of AR and VR markets in recent years but these have not realised yet\footnote{https://www.marketresearchfuture.com/reports/augmented-reality-virtual-reality-market-6884}. This is due to many factors including equipment cost, available content and the physiological effects of `immersion' (particularly over extended time periods) due to conflicting sensory interactions \citep{Ng:study:2020}. 
VR can be used to simulate a real workspace for training workers for the sake of safety and to prevent the real-world consequences of failure \citep{Laver:Virtual:2017}.  { In the healthcare industry, VR is being increasingly used in various sectors, ranging from surgical simulation to physical therapy \citep{Keswani:World:2020}.}

Gaming is often cited as a major market for VR, along with related areas such as pre-visualisation of designs or creative productions (e.g., in building, architecture and filmmaking). {A good list of VR games can be found in many article\footnote{https://www.forbes.com/sites/jessedamiani/2020/01/15/the-top-50-vr-games-of-2019/?sh=42279941322d}. Deep learning technologies have been exploited  in many aspects of gaming, for example in VR/AR game design \citep{Zhang:design:2020} and emotion detection while using VR to improve the user's immersive experience \citep{Quesnel:deep:2018}. More recently AI gaming methods have been extended into the area of virtual production, where the tools are scaled to produce dynamic virtual environments for filmmaking}

AR perhaps has more early potential for growth than VR and uses have been developed in education and to create shared information, work or design spaces,  where it can provide added 3D realism for the users interacting in the space \citep{Palmarini:systematic:2018}.  AR has also gained interest in augmenting experiences in movie and theatre settings\footnote{https://www.factor-tech.com/feature/lifting-the-curtain-on-augmented-reality-how-ar-is-bringing-theatre-into-the-future/}. A review of current and future trends of AR and VR systems can be found in \citep{Bastug:toward:2017}. 

MR combines the real world with digital elements {(or the virtual world) \citep{Milgram:Taxonomy:1994}}. It allows us to interact with objects and environments in both the real and virtual world by using touch technology and other sensory interfaces, {to merge} reality and imagination and to provide more engaging experiences. Examples of MR applications include the `MR Sales Gallery' used by large real estate developers\footnote{https://dynamics.microsoft.com/en-gb/mixed-reality/overview/}. It is  a virtual sample room that simulates the environment for customers to experience the atmosphere of an interactive residential project. The growth of  VR, AR and MR technologies is described by Immerse UK in their recent report on the immersive economy in the UK 2019\footnote{https://www.immerseuk.org/wp-content/uploads/2019/11/The-Immersive-Economy-in-the-UK-Report-2019.pdf}. Extended reality (XR) is a newer technology that combines VR, AR and MR with internet connectivity,  which opens further opportunities across industry, education, defence, health, tourism and entertainment \citep{Chuah:Why:2018}. 

{
An immersive experience with VR or MR requires good quality, high-resolution, animated worlds or 360-degree video content
\citep{Ozcinar:visual:2018}. This poses new problems for data compression and visual quality assessment, which are the subject of increased research activity currently \citep{Xu:360:2020}. }
AI technologies have been employed to make AR/VR/MR/XR content more exciting and realistic, to robustly track and {localize} objects and users in the environment. For example, automatic map reading using image-based {localization}   \citep{Panphattarasap:automated:2018},  and gaze estimation  \citep{Anantrasirichai:fixation:2016, Soccini:gaze:2017}. Oculus Insight, by Facebook, uses visual-inertial SLAM (simultaneous localization and mapping) to generate real-time maps and position tracking\footnote{https://ai.facebook.com/blog/powered-by-ai-oculus-insight/}.
{ More sophisticated approaches, such as Neural Topological SLAM,  leverage semantics and geometric information to improve long-horizon navigation \citep{Chaplot:TopologicalSLAM:2020}.
Combining audio and visual sensors can further improve navigation of egocentric observations in complex 3D environments, which can be done through deep reinforcement learning approach \citep{Chen:SoundSpaces:2020}.}

\subsubsection{Deepfakes}
\label{deepfakes}

Manipulations of visual and auditory media, either for amusement or malicious intent,  are not new. However, advances in AI and ML methods have taken this to another level, improving their realistism and providing automated processes that make them easier to render. Text generator tools, such as those by OpenAI, can generate coherent paragraphs of text with basic comprehension, translation and summarization but have also been used to create fake news or abusive spam on social media\footnote{https://talktotransformer.com/}. Deepfake technologies can also create realistic fake videos by replacing some parts of the media with synthetic content.  For example, substituting someone's face {while} hair, body and action remain the same (Fig. \ref{fig:poseanimation} (b)\footnote{https://www.youtube.com/watch?time\_continue=2\&v=ANXucrz7Hjs \label{fn:derpfakes}}). Early research created mouth movement synthesis tools capable of making the subject appear to say something different from the actual narrative, e.g., President Barack Obama is lip-synchronized to a new audio track in  \citep{Suwajanakorn:obame:2017}. More recently, DeepFaceLab \citep{Perov:DeepFaceLab:2020} provided a state-of-the-art tool for face replacement; however manual editing is still required in order to create the most natural appearance. Whole body movements have been generated via learning from a source video to synthesize the positions of arms, legs and body of the target  \citep{Chan:Everybody:2019}. 

Deep learning approaches to Deepfake generation primarily employ generative neural network architectures, e.g., VAEs \citep{Kietzmann:deepfakes:2020} and GANs \citep{Zakharov:Few:2019}. Despite rapid progress in this area, the creation of perfectly natural figures remains challenging; for example deepfake faces often do not blink naturally. Deepfake techniques have been widely used to create pornographic images of celebrities, to cause political distress or social unrest, for purposes of blackmail and to announce fake terrorism events or other disasters. This has resulted in several countries banning non-consensual deepfake content. To counter these often malicious attacks,  a number of approaches have been reported and introduced to detect fake digital content  \citep{Guera:deepfake:2018,Li:Exposing:2019,Hasan:combating:2019}.

\subsubsection{Content and Captions}

There are many approaches that attempt to interpret an image or video and then automatically generate captions based on its content \citep{Xia:recurrent:2005, Pu:Variational:2016, Xu:Learnging:2017}. This can successfully be achieved through object recognition (see Section \ref{sssec:recog}); YouTube has provided this function for both video-on-demand and livestream videos\footnote{https://support.google.com/youtube/answer/6373554?hl=en}.

The other way around, AI can also help to generate a new image from text. However, this problem is far more complicated; attempts so far have been based on GANs. Early work by \citet{Mansimov:Generating:2016} was capable of generating background image content with relevant colors but with blurred foreground details.  A conditioning augmentation technique was proposed to stabilize the training process of the conditional GAN, and also to improve the diversity of the generated samples  \citep{Zhang:StackGAN:2017}. Recent methods with significantly increased complexity are capable of learning to generate an image in an object-wise fashion, leading to more natural-looking results \citep{Li:Object:2019}. However,  limitations remain, for example artefacts often appear around object boundaries or inappropriate backgrounds can be produced if the words of the caption are not given in the correct order.

\subsection{Information Analysis} 

AI has proven capability to process and adapt to large amounts of training data. It can learn and analyze the characteristics of these data, making it possible to classify content and predict outcomes with high levels of confidence. Example applications include advertising and film analysis, as well as image or video retrieval, for example enabling producers to acquire information, analysts to better market products or journalists to retrieve content relevant to an investigation. 

{
\subsubsection{Text categorization }
\label{sssec:textgen}
}

{Text categorization is a core application of NLP. This generic text processing task is useful in indexing documents for subsequent retrieval and content analysis (e.g., spam detection, sentiment classification, and topic classification). It can be thought of as the generation of summarised texts from full texts. Traditional techniques for both multi-class and multi-label classifications include decision trees, support vector machines \citep{Kowsari:text:2019}, term frequency–inverse document frequency \citep{Azam:comparison:2012},} {and extreme learning machine \citep{Mohammad:Regularizing:2021}. Unsupervised learning with self-organizing maps has also been investigated \citep{Pawar:Comparative:2012}.} 
{Modern NLP techniques are based on deep learning, where generally the first layer is an embedding layer that converts words to vector representations. Additional CNN layers are then added to extract text features and learn word positions \citep{Johnson:Effective:2015}. RNNs (mostly based on LSTM architectures) have also been concatenated to learn sentences and give prediction outputs \citep{Chen:Ensemble:2017, Gunasekara:review:2018}. A category sentence generative adversarial network has also been proposed that combines GAN, RNN and reinforcement learning to enlarge training datasets, which improves performance for sentiment classification \citep{Li:Generative:2018}. Recently, an attention layer has been integrated into the network to provide semantic representations in aspect-based sentiment analysis \citep{Trusca:hybrid:2020}.
The artist, Vibeke Sorensen, has applied AI techniques to categorize  texts  from global social networks such as Twitter into six live emotions and display the `Mood of the Planet' artistically using six different colors\footnote{http://vibeke.info/mood-of-the-planet/}.
}

\subsubsection{Advertisements and Film Analysis}

AI can assist creators in matching content more effectively to their audiences, for example recommending music and movies in a streaming service, like Spotify or Netflix. Learning systems  have also been used to {characterize} and target individual viewers, {optimizing} the time they spend on advertising \citep{Lacerda:Learning:2006}. This approach assesses what users look at and how long they spend browsing adverts, participating on social media platforms. In addition, AI can be used to inform  how adverts should be presented to help boost their effectiveness, for example by identifying suitable customers and showing the ad at the right time. This normally involves gathering and analysing personal data in order to predict preferences  \citep{Golbeck:Predicting:2011}. 

Contextualizing social-media conversations can also help advertisers understand how consumers feel about products and to detect fraudulent ad impressions \citep{Ghani:Social:2019}. This can be achieved using NLP methods \citep{Young:recent:2018}.
Recently, an AI-based data analysis tool has been introduced {to assist filmmaking companies} to develop strategies for how, when and where  prospective films should be released \citep{Dodds:AI:2020}. The tool employs ML approaches to model the patterns of historical data about film performances associating with the film's content and themes. This is also used in gaming industries, where  the behaviour of each player is analyzed so that the company can better understand their style of play and decide when best to approach them to make money\footnote{https://www.bloomberg.com/news/articles/2017-10-23/game-makers-tap-ai-to-profile-each-player-and-keep-them-hooked}.

\subsubsection{Content Retrieval}
\label{sssec:retrieval}
Data retrieval is an important component in many creative processes, since producing a new piece of work generally requires undertaking a significant amount of research at the start. Traditional retrieval technologies employ metadata or annotation text (e.g., titles, captions, tags, keywords and descriptions) to the source content \citep{Jeon:Automatic:2003}. The manual annotation process needed to create this metadata is however very time-consuming.  AI methods have enabled automatic annotation by supporting the analysis of  media based on audio and object recognition and scene understanding \citep{Wu:Deep:2015, Amato:searching:2017}.  

In contrast to traditional concept-based approaches, content-based image retrieval (or query by image content (QBIC)) analyzes the content of an image rather than its metadata. A reverse image search technique (one of the techniques Google Images uses\footnote{https://images.google.com/}) extracts low-level features from an input image, such as points, lines, shapes, colors and textures. The query system then searches for related images by matching these features within the search space. Modern image retrieval methods often employ deep learning techniques, enabling image to image searching by extracting low-level features  and then combining these to form semantic representations of the reference image that can be used as the basis of a search \citep{Wan:DeepRetrieval:2014}. For example, when a user uploads an image of a dog to Google Images, the search engine will return the dog breed, show similar websites by searching with this {keyword}, and also show selected images that are visually similar to that dog, e.g., with similar colors and background. These techniques have been further improved by exploiting features at local, regional and global image levels \citep{Gordo:DeepRetrieval:2016}. GAN approaches are also popular, associated with learning-based hashing which was proposed for scalable image retrieval \citep{Song:Binary:2018}. Video retrieval can be more challenging due to the requirement for understanding activities, interactions between objects and unknown context; RNNs have provided a natural {extension that supports the extraction} of sequential behaviour in this case \citep{Jabeen:Video:2018}.

Music information retrieval extracts features of sound, and then converts these to a meaningful representation suitable for a query engine. Several methods for this have been reported, including automatic tagging, query by humming, search by sound and acoustic fingerprinting \citep{Kaminskas:music:2012}. 

\subsubsection{Recommendation Services}

A recommendation engine is a system that suggests products, services, information to users based on analysis of data. For example, a music curator creates a soundtrack or a playlist that has songs with similar mood and tone, bringing related content to the user. Curation tools, capable of searching large databases and creating recommendation shortlists, have become popular because they can save time, elevate brand visibility and increase connection to the audience. { The techniques used in recommendation systems generally fall into three categories: i) content-based filtering, which uses a single user's data , ii) collaborative filtering,  the most prominent approach, that derives suggestions from many other users,  and iii) knowledge-based system, based on specific queries made by the user, which is generally employed in complex domains, where the first two cannot be applied. The approach can be hybrid; for instance where content-based filtering exploits individual metadata and collaborative filtering finds overlaps between user playlists. } Such systems build a profile of what the users listen to or watch, and then look at what other people who {have similar profiles listen to or watch}. ESPN and Netflix have partnered with Spotify to curate playlists from the documentary `The Last Dance'. Spotify has created music and podcast playlists that viewers can check out after watching the show\footnote{https://open.spotify.com/show/3ViwFAdff2YaXPygfUuv51}. 

Content summarization is a fundamental tool that can support recommendation services. Text categorization approaches extract important content from {a document into key indices} (see Section \ref{sssec:textgen}). RNN-based models incorporating  attention models have been employed to successfully generate a summary in the form of an abstract \citep{rush:nueral:2015}, short paragraph \citep{See:get:2017} or a personalized sentence \citep{Li:Towards:2019}. The gaze behavior of an individual viewer has also been included for personalised text summarization \citep{Yi:Personalized:2020}. The personalized identification of key frames and start points in a video has also been framed as  an optimization problem in  \citep{Chen:Resource:2014}. ML approaches have been developed to perform content-based recommendations. Multimodal features of text, audio, image, and video content  are extracted and used to seek similar content in  \citep{Deldjoo:audio:2018}. This task is relevant to content retrieval, as discussed in Section \ref{sssec:retrieval}. A detailed review of deep learning for recommendation systems can be found in \citep{Batmaz:review:2019}.

\subsubsection{Intelligent Assistants}
Intelligent Assistants employ a combination of AI tools, including many of those mentioned above, in the form of a software agent that can perform tasks or services for an individual. These virtual agents can access information via digital channels to answer questions relating to, for example,  weather forecasts, news items or encyclopaedic enquiries. They can recommend songs, movies and places, as well as suggest routes. They can also manage personal schedules, emails, and reminders. The communication can be in the form of text or voice. The AI technologies behind the intelligent assistants are based on sophisticated ML and NLP methods. Examples of current intelligent assistants include Google Assistant\footnote{https://assistant.google.com/}, Siri\footnote{https://www.apple.com/siri/}, Amazon Alexa and Nina by Nuance\footnote{https://www.nuance.com/omni-channel-customer-engagement/digital/virtual-assistant/nina.html}. Similarly, chatbots and other types of virtual assistants are used for marketing, customer service, finding specific content and information gathering \citep{Xu:Chatbot:2017}.
\subsection{Content Enhancement and Post Production Workflows}
\label{ssec:contentenhance}

{It is often the case that original content (whether images, videos, audio or documents) is not fit for the purpose of its target audience. This could be due to noise caused by sensor limitations,  the conditions prevailing during acquisition, or degradation over time. AI offers the potential to create assistive intelligent tools that improve both quality and management, particularly for mass-produced content. }

\subsubsection{Contrast Enhancement}

The human visual system employs many opponent processes, both in the retina and visual cortex, that rely heavily on differences in color, luminance or motion to trigger salient reactions \citep{David:Intelligent:2021}. Contrast is the difference in luminance and/or color that makes an object distinguishable, and this is an important factor in any subjective evaluation of image quality. Low contrast images exhibit a narrow range of tones and can therefore appear flat or dull. Non-parametric methods for contrast enhancement involve histogram equalisation which spans the intensity of an image between its bit depth limits from 0 to a maximum value (e.g., 255 for 8 bits/pixel). Contrast-limited adaptive histogram equalisation (CLAHE) is one example that { is commonly} used to adjust an histogram and reduce noise amplification \citep{Pizer:adaptive:1987}. Modern methods have further extended performance by exploiting CNNs and autoencoders \citep{Lore:LLNet:2017}, inception modules and residual learning \citep{Tao:LLCNN:2017}. Image Enhancement Conditional Generative Adversarial Networks (IE-CGANs) designed to process both visible and infrared images have been proposed by \citet{Kuang:single:2019}. Contrast enhancement, along with other methods to be discussed later, suffer from a fundamental lack of data for supervised training because real image pairs with low and high contrast are unavailable \citep{Jiang:EnlightenGAN:2021}. Most of these methods therefore train their networks with synthetic data (see Section \ref{ssec:datasets}). 

\subsubsection{Colorization}

Colorization is the process that adds or restores color in visual media. This can be useful in coloring archive black and white content, enhancing infrared imagery (e.g., in low-light natural history filming) and also in restoring the color of aged film.  A good example is the recent film ``They Shall Not Grow Old" (2018) by Peter Jackson, that colorized (and corrected for speed and jerkiness, added sound and converted to 3D) 90 minutes of footage from World War One. The workflow was based on extensive studies of WW1 equipment and uniforms as a reference point and involved a time-consuming use of post production tools. 

The first AI-based techniques for colorization used a CNN with only three convolutional layers to convert a grayscale image into chrominance values and refined them with bilateral filters to generate a natural color image \citep{Cheng:Colorization:2015}. A deeper network, but still only with eight dilated convolutional layers, was proposed  a year later \citep{Zhang:colorful:2016}. This network captured better semantics, resulting in an improvement on images with distinct foreground objects. {Encoder-decoder networks are employed in \citep{ Xu:Stylization:2020}}

\begin{figure*}[t!]
	\centering
	\vspace{5mm}
      		 \includegraphics[width=\textwidth]{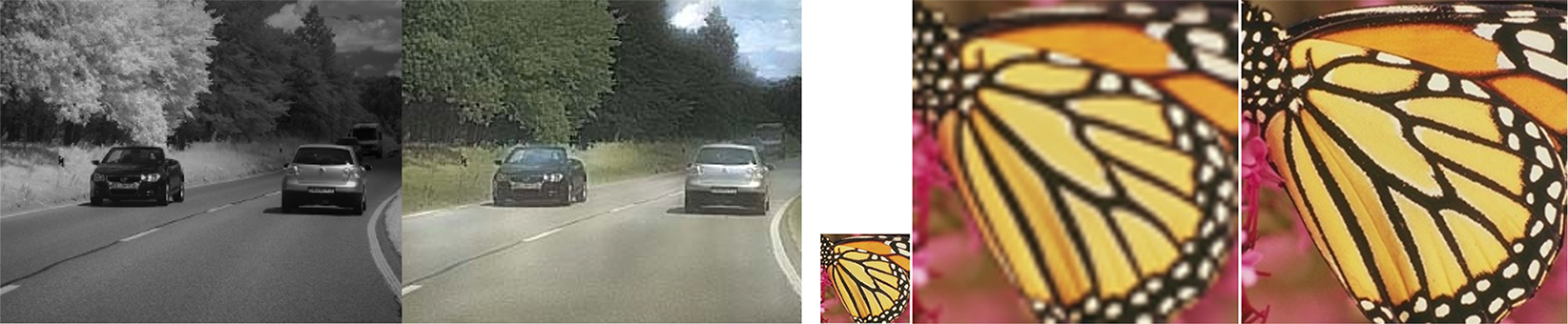}
      		 \begin{scriptsize}
      		 \begin{tabular}{x{2.3cm}x{2.6cm}x{1.2cm}x{2cm}x{1.8cm}}
		Original IR image & Colorized image & Original LR & Bicubic interpolate & SRGAN
\end{tabular}   \end{scriptsize}	
      		 \begin{tabular}{x{5.7cm}x{6.8cm}}
			(a) & (b)
\end{tabular} 
					\caption{\small Image enhancement. (a) Colorization for infrared image  \citep{Limmer:Infrared:2016}. (b) Super-resolution \citep{Ledig:Photo:2017}}
    \label{fig:enhancedimage}
\end{figure*}

Colorization remains  a challenging problem for AI as recognized in the  recent Challenge in Computer Vision and Pattern Recognition Workshops (CVPRW) in 2019 \citep{Nah:NTIRE:2019}. Six teams competed and all of them employed deep learning methods.  Most of the methods adopted an encoder-decoder or a structure based on U-Net \citep{Ronneberger:Unet:2015}. The deep residual net (NesNet) architecture \citep{He:ResNet:2016} and the dense net (DenseNet) architecture \citep{Huang:Densely:2017} have both demonstrated effective conversion of gray scale to natural-looking color images. More complex architectures have been developed based on GAN structures \citep{Zhang:seftattention:2019}, for example DeOldify and NoGAN  \citep{Antic:DeOldify:2020}. The latter model was shown to reduce temporal color flickering on the video sequence, which is a common problem when enhancing colors on an individual frame by frame basis. Infrared images have also been converted to natural color images using CNNs \citep[e.g.,][]{Limmer:Infrared:2016} (Fig. \ref{fig:enhancedimage} (a)) and GANs  \citep[e.g.,][]{Suarez:Infrared:2017,Kuang:Thermal:2020}.

\subsubsection{Upscaling Imagery: Super-resolution Methods}

Super-resolution (SR) approaches have gained popularity in recent years, enabling the upsampling of images and video spatially or temporally. This is useful for up-converting legacy content for compatibility with modern formats and displays. SR methods increase the resolution (or sample rate) of a low-resolution (LR) image (Fig. \ref{fig:enhancedimage} (b)) or video. In the case of video sequences, successive frames can, for example, be employed to construct a single high-resolution (HR) frame. Although the basic concept of the SR algorithm is quite simple, there are many problems related to perceptual quality and restriction of available data. For example, the LR video may be aliased and exhibit sub-pixel shifts between frames and hence some points in the HR frame do not correspond to any information from the LR frames.

With deep learning-based technologies, the LR and HR images are matched and used for training architectures such as  CNNs,  to provide high quality upscaling potentially using only a single LR image \citep{Dong:Learning:2014}.  Sub-pixel convolution layers can be introduced to improve fine details in the image, as reported by \citeauthor{Shi:Real:2016}. Residual learning and generative models are also employed, \citep[e.g.,][]{Kim:Accurate:2016,Tai:SR:2017}. A generative model with a VGG-based\footnote{VGG is a popular CNN, originally developed for object recognition by the Visual Geometry Group at the University of Oxford \citep{Simonyan:VGG:2015}. {See Section \ref{ssec:cnn} for more detail.}}  perceptual loss function has been shown to significantly improve quality and sharpness when used with the SRGAN by  \citet{Ledig:Photo:2017}.   \citet{Wang:Fully:2018} proposed a progressive multi-scale GAN for perceptual enhancement, where pyramidal decomposition is combined with a DenseNet architecture \citep{Huang:Densely:2017}.  The above techniques seek to learn implicit redundancy that is present in natural data to recover missing HR information from a single LR instance. For single image SR, the review by \citeauthor{Yang:Deep:2019} suggests that methods such as EnhanceNet  \citep{Sajjadi:EnhanceNet:2017} and SRGAN \citep{Ledig:Photo:2017}, that achieve high subjective quality with good sharpness and textural detail, cannot simultaneously achieve low distortion loss (e.g., mean absolute error (MAE) or peak signal-to-noise-ratio (PSNR)). A comprehensive survey of image SR is provided by \citet{Wang:Deep:2020}. This observes that more complex networks generally produce better PSNR results and that most state-of-the-art methods are based on residual learning and use $\ell_1$ as one of training losses \citep[e.g.,][]{Zhang:image:2018,Dai:Second:2019}.

When applied to video sequences, super-resolution methods can exploit temporal correlations across frames as well as local spatial correlations within them. Early contributions applying  deep learning to achieve video SR gathered multiple frames into a 3D volume which formed the input to a CNN \citep{Kappeler:video:2016}. Later work exploited temporal correlation via a motion compensation process before concatenating multiple warped frames into a 3D volume \citep{Caballero:realtime:2017} using a recurrent architecture \citep{Huang:Bidirectional:2015}. The framework proposed by \citet{Liu:Learning:2018} upscales each frame before applying another network for motion compensation. The original target frame is fed, along with its neighbouring frames, into intermediate layers of the CNN to perform  inter-frame motion compensation during feature extraction \citep{Haris:Recurrent:2019}. EDVR \citep{Wang:EDVR:2019}, the winner of the NTIRE19 video restoration and enhancement challenges in 2019\footnote{\url{https://data.vision.ee.ethz.ch/cvl/ntire19/}}, employs a deformable convolutional network \citep{Dai:Deformable:2017} to align two successive frames. Deformable convolution is also employed in DNLN (Deformable Non-Local Network) \citep{Wang:SR:2019}. At the time of writing, EDVR \citep{Wang:EDVR:2019} and DNLN \citep{Wang:SR:2019} are reported to outperform other methods for video SR, followed by the method of \citet{Haris:Recurrent:2019}. This suggests that deformable convolution plays an important role in  overcoming inter-frame misalignment, producing sharp textural details.

\subsubsection{Restoration}

\begin{figure*}[t!]
	\centering
	\vspace{5mm}
      		 \includegraphics[width=\textwidth]{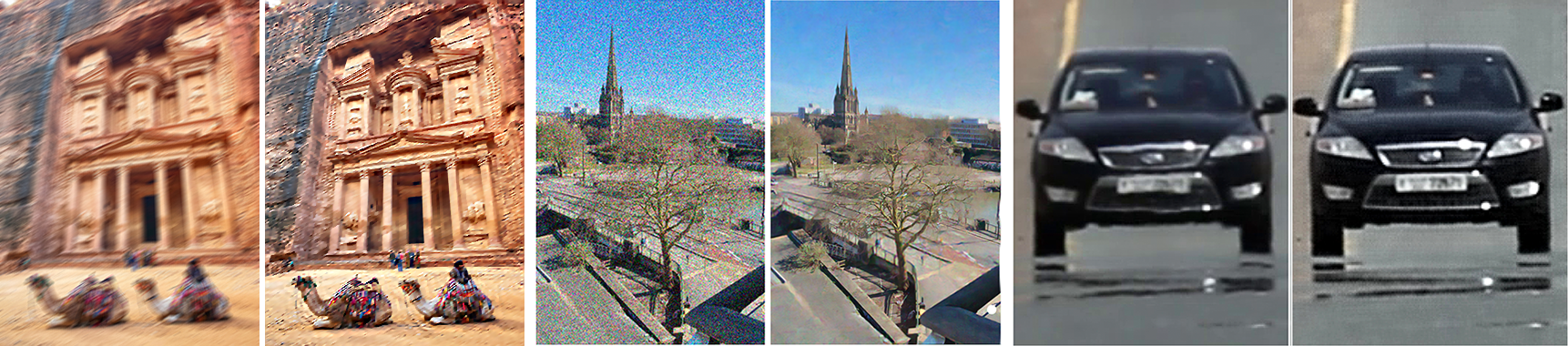}  
      		 \begin{tabular}{x{3.7cm}x{3.4cm}x{3.7cm}}
			(a) & (b) & (c)
\end{tabular} 
					\caption{\small Restoration for (a) deblurring \citep{Zhang:Dynamic:2018}, (b) denoising with DnCNN \citep{Zhang:dncnn:2017}, and (c) turbulence mitigation \citep{Anantrasirichai:Atmospheric:2013}. Left and right  are the original degraded images and the restored images respectively.}
    \label{fig:restoredimage}
\end{figure*}

The quality of a signal can often be reduced due to distortion or damage. This could be due to environmental conditions during acquisition (low light, atmospheric distortions or high motion), sensor characteristics (quantization due to limited resolution or bit-depth or electronic noise in the sensor itself) or ageing of the original medium such as tape of film.  The general degradation model can be written as $I_{obs} = h * I_{ideal} + n$, where $I_{obs}$ is an observed (distorted) version  of the ideal signal $I_{ideal}$,  $h$ is the degradation  operator, $*$ represents convolution, and $n$ is noise. The restoration process tries to reconstruct $I_{ideal}$ from  $I_{obs}$. $h$ and $n$ are values or functions that are dependent on the application. Signal restoration can be addressed as an inverse problem and deep learning techniques have been employed to solve it.  Below we divide restoration into four classes that relate to work in the creative industries with examples illustrated in Fig. \ref{fig:restoredimage}. Further details of deep learning for inverse problem solving can be found in \citep{Lucas:using:2018}.

i) \textbf{Deblurring}:
Images can be distorted by blur, due to poor camera focus or camera or subject motion. Blur-removal is an ill-posed problem represented by a point spread function (PSF) $h$, which is generally unknown. Deblurring methods sharpen an image to increase subjective quality, and also to assist subsequent operations such as optical character recognition (OCR) \citep{Hradis:text:2015} and object detection \citep{Kupyn:DeblurGAN:2018}.  Early work in this area analyzed the statistics of the image and attempted to model physical image and camera properties \citep{Biemond:Iterative:1990}. More sophisticated algorithms such as blind deconvolution (BD), attempt to restore the image and the PSF simultaneously \citep{Jia:single:2007, Krishnan:Blind:2011}. These methods however assume a space-invariant PSF and the process generally involves several iterations. 

As described by the image degradation model, the PSF ($h$) is related to the target image via a convolution operation. CNNs are therefore inherently applicable for solving blur problems \citep{Schuler:Learning:2016}. Deblurring techniques based on CNNs \citep{Nah:Deep:2017} and GANs \citep{Kupyn:DeblurGAN:2018} usually employ residual blocks, where skip connections are inserted every two convolution layers \citep{He:ResNet:2016}.  Deblurring an image from coarse-to-fine scales is proposed in  \citep{Tao:scale:2018}, where the outputs are upscaled and are fed back to the encoder-decoder structure. The high-level features of each iteration are linked in a recurrent manner, leading to a recursive process of learning sharp images from blurred ones. Nested skip connections were introduced by \citet{Gao:dynamic:2019}, where feature maps from multiple convolution layers are merged before applying them to the next convolution layer (in contrast to  the residual block approach where one feature map is merged at the next input). This more complicated architecture improves information flow and results in  sharper images with fewer ghosting artefacts compared to previous methods. 

In the case of video sequences, deblurring can benefit from the abundant information present across neighbouring frames. The DeBlurNet model  \citep{Su:Deep:2017} takes a stack of nearby frames as input and uses synthetic motion blur to generate a training dataset. A Spatio-temporal recurrent network exploiting a dynamic temporal blending network is proposed by \citet{Kim:Online:2017}. \citet{Zhang:Dynamic:2018} have concatenated an encoder, recurrent network and decoder to mitigate motion blur. Recently a recurrent network with iterative updating of the hidden state was trained using a regularization process to create sharp images with fewer ringing artefacts \citep{Nah:Recurrent:2019}, denoted as IFI-RNN. A Spatio-Temporal Filter Adaptive Network (STFAN) has been proposed \citep{Zhou:Spatio:2019}, where the convolutional kernel is acquired from the feature values in a spatially varying manner. IFI-RNN and STFAN produce comparable results and hitherto achieve the best performances in terms of both subjective and objective quality measurements (the average PSNRs of both methods are higher than that of \citep{Kim:Online:2017} by up to 3~dB).

 ii) \textbf{Denoising}:
Noise can be introduced from various sources during signal acquisition, recording and processing, and is normally attributed to sensor limitations when operating under extreme conditions. It is generally {characterize}d in terms of whether it is additive, multiplicative, impulsive or signal dependent, and in terms of its statistical properties.
Not only visually distracting, but noise can also affect the performance of detection, classification and tracking tools. Denoising nodes are therefore commonplace in post production workflows, especially for challenging low light natural history content \citep{Anantrasirichai:Encoding:2020}. In addition, noise can reduce the efficiency of  video compression algorithms, since  the encoder allocates wasted bits to represent  noise rather than signal, especially at low compression levels. This is the reason that film-grain noise suppression tools are employed in certain modern video codecs (Such as AV1) prior to encoding by  streaming and broadcasting organisations. 

The simplest noise reduction technique is weighted averaging,  performed spatially and/or temporally as a sliding window, also known as a moving average filter~\citep{Yahya:video:2016}. More sophisticated methods however perform significantly better and are able to adapt to change noise statistics. These include adaptive spatio-temporal smoothing through anisotropic filtering~\citep{Malm:adaptive:2007}, nonlocal transform-domain group filtering~\citep{Maggioni:BM4D:2012}, Kalman-bilateral mixture model~\citep{Zuo:video:2013}, and spatio-temporal patch-based filtering~\citep{Buades:CFA:2019}. Prior to the introduction of deep neural network denoising, methods such as BM3D (block matching 3-D) \citep{Dabov:BM3D:2007} represented the state of the art in denoising performance.

Recent advances in denoising have almost entirely been based on deep learning approaches and these now represent the state of the art.  RNNs have been employed successfully to remove noise in audio \citep{Maas:Recurrent:2012, Zhang:Deep:2018}. A residual noise map is estimated in the Denoising Convolutional Neural Network (DnCNN) method~\citep{Zhang:dncnn:2017} for image based denoising, and for video based denoising, a spatial and temporal network are concatenated \citep{Claus:ViDeNN:2019} where the latter handles brightness changes and temporal inconsistencies.  FFDNet is a modified form of DnCNN that works on reversibly downsampled subimages \citep{Zhang:FFDNet:2018}. \citet{Liu:Multi:2018} developed MWCNN; a similar system that integrates multiscale wavelet transforms within the network to replace max pooling layers in order to better retain visual information. This integrated a wavelet / CNN denoising system and currently provides the state-of-the-art performance for Additive Gaussian White Noise (AGWN). VNLnet combines a non-local patch search module with DnCNN. The first part extracts features, while the latter mitigates the remaining noise~\citep{Davy:nonlocal:2019}. \citet{Zhao:simple:2019}  proposed a simple and shallow network, SDNet, uses six convolution layers with some skip connection to create a hierarchy of  residual blocks. TOFlow \citep{Xue:Video:2019}  offers an end-to-end trainable convolutional network that performs motion analysis and video processing simultaneously. GANs have been employed to estimate a noise distribution which is subsequently used to augment clean data for training CNN-based denoising networks (such as DnCNN) \citep{Chen:image:2018}. GANs for denoising data have been proposed for medical imaging \citep{Yang:Low:2018}, but they are not popular in the natural image domain due to the limited data resolution of current GANs. However, CycleGAN has recently been modified to attempt denoising and enhancing low-light ultra-high-definition (UHD) videos using a patch-based strategy \citep{Anantrasirichai:Contextual:2021}.

Recently, the Noise2Noise algorithm has shown that it is possible to train a denoising network without clean data, under the assumption that the data is corrupted by zero-mean noise \citep{Lehtinen:noise2noise:2018}. The training pair of input and output images are both noisy and the network learns to minimize the loss function by solving the point estimation problem separately for each input sample. However, this algorithm is sensitive to the loss function used, which can significantly influence the performance of the model. Another algorithm, Noise2Void \citep{Krull:Noise2Void:2019}, employs a novel blind-spot network that does not include the current pixel in the convolution. The network is trained using the noisy patches as input and output within the same noisy patch. It achieves comparable performance to Noise2Noise but allows the network to learn noise characteristics in a single image. 

NTIRE 2020 held a denoising grand challenge within the IEEE CVPR conference that compared many contemporary high performing ML denoising methods on real images \citep{Abdelhamed:NTIRE:2020}. The best competing teams employed a variety of techniques using variants on CNN architectures such as U-Net  \citep{Ronneberger:Unet:2015},  ResNet \citep{He:ResNet:2016} and DenseNet \citep{Huang:Densely:2017}, together with  $\ell_1$ loss functions and ensemble processing including flips and rotations. The survey by \citet{Tian:Deep:2020} states that SDNet \citep{Zhao:simple:2019} achieves the best results on ISO noise, and FFDNet \citep{Zhang:FFDNet:2018} offers the best denoising performance overall, including Gaussian noise and spatially variant noise (non-uniform noise levels). 

Neural networks have also been used for other aspects of image denoising: \citet{Chen:Learning:2018}  have developed specific low light denoising methods using CNN-based methods; \citet{Lempitsky:Deep:2018} have developed a deep learning prior that can  be used to denoise images without access to training data; and \citet{Brooks:Unprocessing:2019} have developed specific neural networks to denoise real images through `unprocessing', i.e. they re-generate raw captured images by inverting the processing stages in a camera to form a supervised training system for raw images. 

iii) \textbf{Dehazing}:
In certain situations, fog, haze, smoke and mist can create mood in an image or video. In other cases, they are considered as distortions that reduce contrast, increase brightness and lower color fidelity. Further problems can be caused by condensation forming on the camera lens. The degradation model can be represented  as: $I_{obs} = I_{ideal} t + A (1-t)$ where $A$ is  atmospheric light and   $t$ is medium transmission. The transmission $t$ is estimated using a dark channel prior based on the observation that the lowest value of each color channel of haze-free images is close to zero \citep{He:single:2011}. In  \citep{Berman:Non:2016}, the true colors are recovered based on the assumption that an image can be faithfully represented with just a few hundred distinct colors. The authors showed that   tight color clusters change because of haze and form lines in RGB space enabling them to be readjusted. The scene radiance ($I_{ideal}$) is attenuated exponentially with depth so some work has included an estimate of the depth map  corresponding to each pixel in the image \citep{Kopf:Deep:2008}. CNNs are  employed to estimate transmission $t$ and dark channel by \citet{Yang:Proximal:2018}. Cycle-Dehazing \citep{Engin:cycle:2018} is used to enhance GAN architecture in CycleGAN \citep{Zhu:Unpaired:2017}. This formulation combines cycle-consistency loss (see Section \ref{sssec:imggen}) and perceptual loss (see Section \ref{ssec:cnn}) in order to improve the quality of textural information recovery and generate visually better haze-free images \citep{Engin:cycle:2018}. A comprehensive study and an evaluation of existing single-image dehazing CNN-based algorithms are reported by \citet{Li:Benchmarking:2019}. It concludes that DehazeNet \citep{Cai:DehazeNet:2016} performs best in terms of perceptual loss, MSCNN \citep{Tang:single:2019} offers the best subjective quality and superior detection performance on real hazy images, and AOD-Net \citep{Li:AOD:2017} is the most efficient.

A related application is underwater photography \citep{Li:Underwater:2016} as commonly used in natural history filmmaking. CNNs are employed to estimate the corresponding transmission map or ambient light of an underwater hazy image in  \citep{Shin:estimate:2016}. More complicated structures merging U-Net, multi-scale estimation, and incorporating cross layer connections to produce even better results are reported by \citet{Hu:Underwater:2018}.

 iv) \textbf{Mitigating atmospheric turbulence}:
When the temperature difference between the ground and the air increases, the air layers move upwards rapidly, leading to  a change in the interference pattern of the light refraction. This is generally observed as a combination of blur, ripple and intensity fluctuations in the scene. Restoring a scene distorted by atmospheric turbulence is a challenging problem. The effect, which  is caused by random, spatially varying, perturbations, makes a model-based solution difficult and, in most cases, impractical. Traditional methods have involved frame selection, image registration, image fusion, phase alignment and image deblurring \citep{Anantrasirichai:Atmospheric:2013, Zhu:Removing:2013, Xie:Removing:2016}. Removing the turbulence distortion from a video containing moving objects is very challenging, as generally multiple frames are used and they are needed to be aligned. Temporal filtering with local weights determined from optical flow is employed to address this by \citet{Anantrasirichai:Atmospheric:2018}. However, artefacts in the transition areas between foreground and background regions can remain. Removing atmospheric turbulence based on  single image processing is proposed using ML by \citet{Gao:Atmospheric:2019}. Deep learning techniques to solve this problem are still in their early stages. However, one method reported employs a CNN to support deblurring \citep{Nieuwenhuizen:deep:2019} and another employs multiple frames using a GAN architecture \citep{Chak:Subsampled:2018}. This however appears only to work well for static scenes.

\subsubsection{Inpainting}

\begin{figure*}[t!]
	\centering
	\vspace{5mm}
      		 \includegraphics[width=\textwidth]{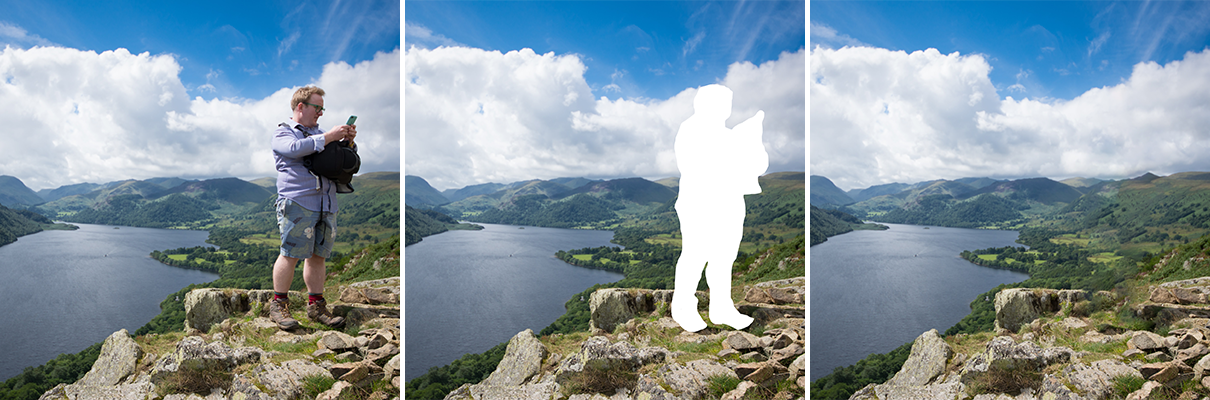}  
					\caption{\small Example of inpainting, (left-right) original image, masking and inpainted image.}
    \label{fig:freeforminpaint}
\end{figure*}

Inpainting is the process of estimating lost or damaged parts of an image or a video.
Example applications for this approach include  the repair of damage caused by cracks, scratches, dust or spots on film or chemical damage resulting in image degradation. Similar problems arise due to data loss during transmission across  packet networks.  Related applications include the removal of unwanted foreground objects or regions of an image and video; in this case the occluded background that is revealed must be estimated. An example of inpainting is shown in Fig. \ref{fig:freeforminpaint}. In digital photography and video editing, perhaps the most widely used tool is Adobe Photoshop\footnote{https://www.adobe.com/products/photoshop/content-aware-fill.html}, where  inpainting is achieved using content-aware interpolation by analysing the entire image to find the best detail to intelligently replace the damaged area.

Recently AI technologies have been reported that model the missing parts of an image using content in proximity to the damage, as well as global information to assist extracting semantic meaning. \citet{Xie:Image:2012} combine sparse coding with deep neural networks pre-trained with denoising auto-encoders. Dilated convolutions are employed in two concatenated networks for spatial reconstruction in the coarse and fine details \citep{Yu:Generative:2018}. Some methods allow users to interact with the process, for example inputting information such as strong edges to guide the solution and produce better results. An example of this image inpainting with user-guided free-form is given by \citet{Yu:Freeform:2019}. Gated convolution is used to learn the soft mask automatically from the data and the content is then generated using both low-level features and extracted semantic meaning. \citet{Chang:Free:2019} extend the work by \citet{Yu:Freeform:2019} to video sequences using a GAN architecture. Video Inpainting, VINet, as reported by  \citet{Kim:Deep:2019} offers the ability to remove moving objects and replace them with content aggregated from both spatial and temporal information using CNNs and recurrent feedback. \citet{Black:Evaluation:2020} evaluated state-of-the-art methods by comparing performance based on the classification and retrieval of fixed images. They reported that DFNet \citep{Hong:Deep:2019}, based on U-Net \citep{Ronneberger:Unet:2015} adding fusion blocks in the decoding layers, outperformed other methods over a wide range of missing pixels.

\subsubsection{Visual Special Effects (VFX)}
Closely related to animation, the use of ML-based AI in VFX has increased rapidly in recent years. Examples include BBC’s His Dark Materials and Avengers Endgame (Marvel)\footnote{https://blogs.nvidia.com/blog/2020/02/07/ai-vfx-oscars/}. 
These both use a combination of physics models with data driven results from AI algorithms to create high fidelity and photorealistic 3D animations, simulations and renderings. ML-based tools transform the actor's face into the film's character using head-mounted cameras and facial tracking markers. With ML-based AI, a single image can be turned into a photorealistic and fully-clothed production-level 3D avatar in real-time  \citep{Hu:Avatar:2017}. Other techniques related to VFX can be found in Section \ref{ssec:contentcreation} (e.g., style transfer and deepfakes), Section \ref{ssec:contentenhance} (e.g., colorization and super-resolution) and Section \ref{ssec:infoextract} (e.g tracking and 3D rendering). AI techniques\footnote{https://www.vfxvoice.com/the-new-artificial-intelligence-frontier-of-vfx/} are increasingly being employed to reduce the human resources needed for certain labour-intensive or repetitive tasks such as match-move, tracking, rotoscoping, compositing and animation \citep{Barber:camera:2016, Torrejon:rotoscope:2020}.

\subsection{Information Extraction and Enhancement}
\label{ssec:infoextract}

AI methods based on deep learning have demonstrated significant success in recognizing and  extracting information from data. They are well suited to this task since successive convolutional layers efficiently perform statistical analysis from low to high level, progressively abstracting meaningful and representative features. Once information is extracted from a signal, it is frequently desirable to enhance it or transform it in some way. This may, for example, make an image more readily interpretable through modality fusion, or translate actions from a real animal to an animation.  This section investigates how AI methods can utilize  explicit information extracted from images and videos to construct such information and reuse it in new directions or new forms.

{\subsubsection{Segmentation}}
\label{sssec:seg}

Segmentation methods are widely employed to partition a signal (typically an image or video) into a form that is semantically more meaningful and easier to analyze or track. The resulting segmentation map indicates the locations and boundaries of semantic objects or regions with parametric homogeneity in an image. Pixels within a region could therefore represent an identifiable object and/or have shared characteristics, such as color, intensity, and texture. Segmentation boundaries indicate the shape of objects and this, together with other parameters,  can be used to identify what the object is. Segmentation can be used as a tool in the creative process, for example assisting with rotoscoping,  masking, cropping and for merging  objects from different sources into a new picture. Segmentation, in the case of video content, also enables the user to change the object or region's characteristics over time, for example through blurring,  color grading or  replacement\footnote{https://support.zoom.us/hc/en-us/articles/210707503-Virtual-Background}.

\begin{figure*}[t!]
	\centering
	\vspace{5mm}
      		 \includegraphics[width=\textwidth]{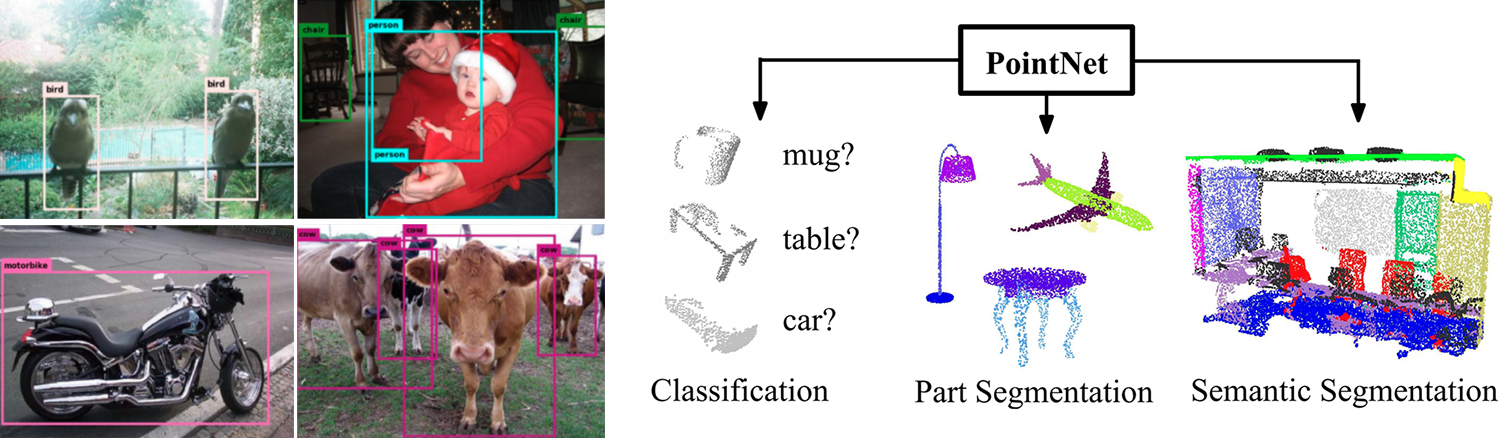}  
      		\begin{tabular}{x{4.25cm}x{7cm}}
			(a) & (b)
\end{tabular} 
					\caption{\small Segmentation and recognition. (a) Object recognition \citep{Kim:Learning:2020}. (b) 3D semantic segmentation \citep{Qi:PointNet:2017}}
    \label{fig:segrecog}
\end{figure*}

Classification systems can be built on top of segmentation in order to detect or identify objects in a scene (Fig. \ref{fig:segrecog}a). This can be compared with the way that humans view a photograph or video, to spot people or other objects, to interpret visual details or to interpret the scene. Since different objects or regions will differ to some degree in terms of the parameters that {characterize} them, we can  train a machine to perform a similar process, providing an understanding of what the image or video contains and activities in the scene. This can in turn support classification, cataloguing and data retrieval. Semantic segmentation classifies all pixels in an image into predefined categories, implying that it processes segmentation and classification simultaneously. { The first deep learning approach to  semantic segmentation employed a fully convolutional network \citep{Long:fully:2015}.  In the same year, the encoder-decoder model in  \citep{ Noh:Learning:2015} and the U-Net architecture \citep{Ronneberger:Unet:2015} were introduced. Following these,  a number of modified networks based on them architectures have been reported  \citep{Taghanaki:semantic:2021}. GANs have also been employed for the purpose of image translation, in this case to translate  a natural image into a segmentation map \citep{Isola:Image:2017}.} The semantic segmentation approach has also been applied to point cloud data to classify and segment 3D scenes, e.g., Fig. \ref{fig:segrecog}b \citep{Qi:PointNet:2017}.

{\subsubsection{Recognition}}
\label{sssec:recog}

Object recognition has been one of the most common targets for AI in recent years, driven by the complexity of the task but also by the huge amount of labeled imagery available for training deep networks. The performance in terms of mean Average Precision (mAP) for detecting 200 classes has increased more than 300\% over the last 5 years \citep{Ciaparrone:Deep:2020}. The Mask R-CNN approach \citep{He:Mask:2017}  has gained popularity due to its ability to separate different objects in an image or a video giving their bounding boxes, classes and pixel-level masks, as demonstrated by \citet{Ren:FasterRCNN:2017}.  Feature Pyramid Network (FPN) is also a popular backbone for object detection \citep{Lin:Feature:2017}.  An in-depth review of object recognition using deep learning can be found in \citep{Zhao:Object:2019} and \citep{Ciaparrone:Deep:2020}. 

YOLO and its variants represent the current state of the art in real-time object detection and tracking \citep{Redmon:YOLO:2016}. A state-of-the-art, real-time object detection system, You Only Look Once (YOLO), works on a frame-by-frame basis and is fast enough to process at typical video rates (currently reported up to 55 fps). YOLO divides an image into regions and predicts bounding boxes using a multi-scale approach and gives probabilities for each region. The latest model, YOLOv4, \citep{Bochkovskiy:YOLOv4:2020}, concatenates YOLOv3 \citep{Redmon:YOLOv3:2018} with a CNN that is 53 layers deep, with SPP-blocks \citep{He:Spatial:2015} or SAM-blocks \citep{Woo:Block:2018} and a multi-scale CNN backbone. YOLOv4 offers real-time computation and high precision (up to 66  mAP on Microsoft's COCO object dataset \citep{Tsung:MSCOCO:2014}). 

On the PASCAL Visual Object Classes (VOC) Challenge datasets \citep{Everingham:voc2012},  YOLOv3  is the leader of object detection on the VOC2010 dataset a with mAP of 80.8\% (YOLOv4 performance on this dataset had not  been reported at the time of writing) and NAS-Yolo is the best for VOC2012 dataset with a mAP of 86.5\%\footnote{http://host.robots.ox.ac.uk:8080/leaderboard/main\_bootstrap.php} (the VOC2012 dataset has a larger number of segmentations than VOC2010). NAS-Yolo \citep{Yang:NAS:2020} employs Neural Architecture Search (NAS) and reinforcement learning to find the best augmentation policies for the target. In the PASCAL VOC Challenge for semantic segmentation, FlatteNet \citep{Cai:Flattenet:2019} and FDNet \citep{Zhen:Learning:2019} lead the field achieving the mAP of 84.3\% and 84.0\% on VOC2012 data, respectively. FlatteNet integrates fully convolutional network with pixel-wise visual descriptors converting from feature maps.  FDNet links all feature maps from the encoder to each input of the decoder leading to really dense network and precise segmentation. 
On the Microsoft COCO object dataset,  MegDetV2 \citep{Li:MegDetV2:2019} ranks first on both the detection leaderboard and the semantic segmentation leaderboard. MegDetV2 combines ResNet with FPN and uses deformable convolution to train the end-to-end network with large mini-batches. 

Recognition of speech and music has also been successfully achieved using deep learning methods. Mobile phone apps that capture a few seconds of sound or music, such as Shazam\footnote{https://www.shazam.com/gb/company}, {characterize} songs based on an audio fingerprint using a spectrogram (a time-frequency graph) that is used to search for a matching fingerprint in a database. Houndify by SoundHound\footnote{https://www.soundhound.com/} exploits speech recognition and searches content across the internet. This technology also provides voice interaction for {in-car systems}. Google proposed a full visual-speech recognition system that maps videos of lips to sequences of words using spatiotemporal CNNs and LSTMs \citep{Shillingford:Large:2019}.

Emotion recognition has also been studied for over a decade. AI methods have been used to learn, interpret and respond to human emotion,  via speech (e.g., tone, loudness, and tempo) \citep{Kwon:Emotion:2003}, face detection (e.g., eyebrows, the tip of nose, the corners of mouth) \citep{Ko:brief:2018}, and both audio and video \citep{Shamim:emotion:2019}. Such systems have also been used in security systems and for fraud detection.

A further task, relevant to video content, is action recognition. This involves capturing spatio-temporal context across frames, for example:  jumping into a pool, swimming, getting out of the pool.  Deep learning has again been extensively exploited in this area{, with  the first report based on a 3D CNN \citep{Ji:3D:2013}.} An excellent {state-of-the-art review} on action recognition can be found in \citep{Yao:review:2019}). More recent advances include temporal segment networks \citep{Wang:TSN:2016} and temporal binding networks, where the fusion of audio and visual information is employed \citep{Kazakos:EPIC:2019}. {  EPIC-KITCHENS, is a large dataset focused on egocentric vision that provides audio-visual, non-scripted recordings in native environments \citep{Damen:EPICKITCHEN:2018}; it has been extensively used to train action recognition systems.
Research on sign language recognition is also related to creative applications, since it studies body posture, hand gesture, and face expression, and hence involves segmentation, detection, classification and 3D reconstruction \citep{Jalal:American:2018, Adithya:deep:2020, Kratimenos:3D:2020}. Moreover, visual and linguistic modelling has been combined to enable translation between spoken/written language and continuous sign language videos \citep{Bragg:sign:2019}.
}

{
\subsubsection{Salient object detection}

Salient object detection (SOD) is a task based on visual attention mechanisms, in which algorithms aim to identify objects or regions that are likely to be the focus of attention. SOD methods can benefit the creative industries in applications such as image editing \citep{Cheng:RepFinder:2010, Mejjati:Parametric:2020}, content interpretation \citep{Rutishauser:is:2004}, egocentric vision \citep{Anantrasirichai:Fixation:2018}, VR \citep{Ozcinar:visual:2018},  and compression \citep{Gupta:visual:2013}. The purpose of SOD differs from fixation detection, which predicts where humans look, but there is a strong correlation between the two \citep{Borji:salient:2019}. In general, the SOD process involves two tasks: saliency prediction and segmentation.  Recent supervised learning technologies have significantly improved the performance of SOD. \citet{Hou:Deeply:2019} merge multi-level features of a VGG network with fusion and cross-entropy losses. A survey by \citet{Wang:salient:2021} reveals that most SOD models employ VGG and ResNet as backbone architectures and train the model with the standard binary cross-entropy loss. More recent work has developed the end-to-end framework with GANs \citep{Wang:SaliencyGAN:2020} and some works include depth information from RGB-D cameras \citep{Jiang:cmSalGAN:2020}. More details on the recent SOD on RGB-D data can be found in  \citep{Zhou:RGBD:2021}. When detecting salient objects in the video, an LSTM module is used to learn saliency shifts \citep{Fan:shifting:2019}. The SOD approach has also been extended to co-salient object detection (CoSOD), aiming to detect the co-occurring salient objects in multiple images \citep{Fan:taking:2020}.
}

\subsubsection{Tracking}

Object tracking is the temporal process of locating objects in consecutive video frames. It takes an initial set of object detections (see Section \ref{sssec:recog}), creates a unique ID for each of these initial detections, and then tracks each of the objects, via their properties, over time. Similar to segmentation, object tracking can support the creative process, particularly in editing. For example, a user can identify and edit a particular area or object in one frame and, by tracking the region, these adjusted parameters can be applied to the rest of the sequence regardless of object motion. Semi-supervised learning is also employed in SiamMask \citep{Wang:Fast:2019} offering the user an interface to define the object of interest and to track it over time.  

Similar to object recognition, deep learning has become an effective tool for object tracking, particularly when tracking multiple objects in the video \citep{Ciaparrone:Deep:2020}. Recurrent networks have been integrated with object recognition methods to track the detected objects over time  \citep[e.g.,][]{Fang:Track:2016,Milan:Online:2017,Gordon:Re3:2018}. 
VOT benchmarks \citep{Kristan:VOT:2016} have been reported for real-time visual object tracking challenges run in both ICCV and ECCV conferences, and the performance of tracking has been observed to improve year on year. The best performing methods include Re$^3$ \citep{Gordon:Re3:2018} and Siamese-RPN \citep{Li:High:2018} achieving 150 and 160 fps at the expected overlap of 0.2, respectively. MOTChallenge\footnote{https://motchallenge.net/} and KITTI\footnote{http://www.cvlibs.net/datasets/kitti/eval\_tracking.php}  are the most commonly used datasets for training and testing multiple object tracking (MOT). At the time of publishing, ReMOTS \citep{Yang:ReMOTS:2020} is currently the best performer  with a mask-based MOT accuracy of 83.9\%. ReMOTS fuses the segmentation results of the Mask R-CNN \citep{He:Mask:2017} and  a ResNet-101 \citep{He:ResNet:2016} backbone extended with FPN.
	

\subsubsection{Image Fusion}
\label{sssec:fusion}

\begin{figure*}[t!]
	\centering
	\vspace{5mm}
      		 \includegraphics[width=\textwidth]{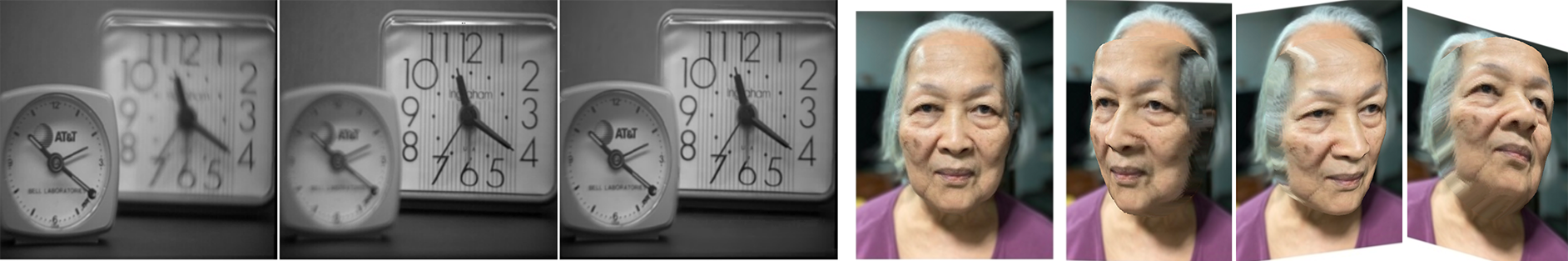}  
      		 \begin{scriptsize}
      		  \begin{tabular}{x{1.8cm}x{1.8cm}x{1.8cm}x{1.7cm}x{3.2cm}}
			 Original image 1 & Original image 2 & Fused image & Original 2D face & Reconstructed 3D face
\end{tabular} 
      		 \end{scriptsize}
      		 \begin{tabular}{x{6cm}x{6cm}}
			(a) & (b)
\end{tabular} 
					\caption{\small Information Enhancement. (a) Multifocal image fusion. (b) 2D to 3D face conversion generated using the algorithm proposed by \citet{Jackson:3Dface:2017}}
    \label{fig:infoenhance}
\end{figure*}

Image fusion provides a mechanism to combine multiple images (or regions therein, or their associated information) into a single representation that has the potential to aid human visual perception and/or subsequent image processing tasks. A fused image (e.g., a combination of IR and visible images) aims to express the salient information from each source image without introducing artefacts or inconsistencies. A number of applications have exploited  image fusion to combine complementary information into a single image, where the capability of a single sensor is limited by design or observational constraints. Existing pixel-level fusion schemes range from simple averaging of the pixel values of registered (aligned) images to more complex multiresolution pyramids, sparse methods \citep{Anantrasirichai:Image:2020} and methods based on complex wavelets \citep{Lewis:pixel:2007}. Deep learning techniques have been successfully employed in many image fusion applications. An all-in-focus image is created using multiple images of the same scene taken with different focal settings \citep{Liu:multi:2017} (Fig. \ref{fig:infoenhance} (a)). Multi-exposure deep fusion is used to create high-dynamic range images by \citet{Prabhakar:DeepFuse:2017}. A review of deep learning for pixel-level image fusion can be found in \citep{Liu:Deep:2018}. Recently, GANs have also been developed for this application \citep[e.g.,][]{Ma:FusionGAN:2019}, with an example of image blending using a guided mask \citep[e.g.,][]{Wu:GPGAN:2019}. 

The performance of a fusion algorithm is difficult to quantitatively assess as no ground truth exists in the fused domain. \citet{Ma:Infrared:2019} shows that a guided filtering-based fusion \citep{Li:Image:2013} achieves the best results based on the visual information fidelity (VIF) metric, but proposed that fused images with very low correlation coefficients, measuring  the degree of linear correlation between the fused image its source images, also works well compared to subjective assessment.

\subsubsection{3D Reconstruction and Rendering}
\label{sssec:3Dreconstruct}

In the human visual system, a stereopsis process (together with many other visual cues and priors \citep{David:Intelligent:2021} creates a perception of three-dimensional (3D) depth from the combination of  two spatially separated signals received by the visual cortex from our retinas. The fusion of these two slightly different pictures gives the sensation of strong three-dimensionality by matching similarities. To provide stereopsis in machine vision applications, images are captured simultaneously from two cameras with parallel camera geometry, and an implicit geometric process is used to extract 3D information from these images. This process can be extended using multiple cameras in an array to create a full volumetric representation of an object. This approach is becoming increasingly popular in the creative industries, especially for special effects that create digital humans\footnote{https://www.dimensionstudio.co/solutions/digital-humans} in high end movies or live performance. 

To convert 2D to 3D representations  (including 2D+t to 3D), the first step is normally depth estimation, which is performed using stereo or multi-view RGB camera arrays. Consumer RGB-D sensors can also be used for this purpose \citep{Maier:Intrinsic3D:2017}. Depth estimation, based on disparity can also be assisted by motion parallax (using a single moving camera),  focus, and perspective. For example, motion parallax is learned using a chain of encoder-decoder networks by \citet{Ummenhofer:DeMoN:2017}. Google Earth has computed topographical information from images captured using aircraft and added texture to create a 3D mesh. As the demands for higher depth accuracy have increased and  real-time computation has become feasible, deep learning methods (particularly CNNs)  have gained more attention. A number of network architectures have been proposed for stereo configurations, including a pyramid stereo matching network (PSMNet) \citep{Chang:pyramid:2018}, a stacked hourglass architecture \citep{Newell:hourglass:2016}, a sparse cost volume network (SCV-Net)  \citep{Lu:sparse:2018}, a fast densenet \citep{Anantrasirichai:Fast:2020} and a guided aggregation net (GA-Net) \citep{Zhang:GANet:2019}. On the KITTI Stereo dataset benchmark  \citep{Geiger:Are:2012}, the team, called LEAStereo from Monash University, ranks 1st at the time of writing (the number of erroneous pixels reported as 1.65\%). They exploit neural architecture search (NAS) technique\footnote{\url{http://www.cvlibs.net/datasets/kitti/eval_scene_flow.php?benchmark=stereo}} to build the best network designed by another neural network. 


3D reconstruction is generally divided into: volumetric, surface-based, and multi-plane representations. Volumetric representations can be achieved by  extending the 2D convolutions used in image analysis. Surface-based representations, e.g., meshes, can be more memory-efficient, but are not regular structures and thus do not easily map onto deep learning architectures.  The state-of-the-art methods for volumetric and surface-based representations are Pix2Vox \citep{Xie:Pix2Vox:2019} and AllVPNet \citep{Soltani:Synthesizing:2017} reporting an Intersection-over-Union (IoU) measure of 0.71 and 0.83 constructed from 20 views on the ShapeNet dataset benchmark \citep{Chang:ShapeNet:2015}). {GAN architectures have been used to generate non-rigid surfaces from a monocular image \citep{Shimada:IsMo:2019}.}
The third type of representation is formed from  multiple planes of the scene. It is a trade-off between the first two representations -- efficient storage  and amenable to training with deep learning. The method in  \citep{Flynn:DeepView:2019}, developed by Google, achieves view synthesis with learned gradient descent. A review of state-of-the-art 3D reconstruction from images using deep learning can be found in \citep{Han:Image:2019}. 

{Recently, low-cost video plus depth (RGB-D) sensors have become widely available. Key challenges related to  RGB-D video processing have included  synchronisation, alignment and data fusion between {multimodal} sensors \citep{Malleson:3D:2019}.  Deep learning approaches have also been used to achieve  semantic segmentation, multi-model feature matching and noise reduction for RGB-D information \citep{Zollhofer:RGBD:2018}. Light field cameras, that capture the intensity and direction of light rays, produce denser data than the RGB-D cameras. Depth information of a scene can be extracted from the displacement of the image array, and 3D rendering has been reported using deep learning approaches in  \citep{Shi:learning:2020}. Recent state-of-the-art light field methods can be found in the review by \citet{Jiang:light:2020}.}

3D reconstruction from a single image is an ill-posed problem. However, it is possible with deep learning due to the network's ability to learn semantic meaning (similar to object recognition, described in Section \ref{sssec:recog}). Using  a 2D RGB training image with 3D ground truth, the model can predict what kind of scene and objects are contained in the test image. Deep learning-based methods also provide state-of-the-art performance for generating the corresponding right view from a left view in a stereo pair  \citep{Xie:Deep3D:2016}, and for converting  2D face images to 3D face reconstructions { using CNN-based encoder-decoder architectures   \citep{Bulat:How:2017, Jackson:3Dface:2017}, autoencoders \citep{Tewari:High:2020} and GANs \citep{Tian:CRGAN:2018}} (Fig. \ref{fig:infoenhance} (b)).  Creating 3D models of bodies from photographs is the focus of \citep{Kanazawa:End:2018}. Here, a CNN is used to translate a single 2D image of a person into parameters of shape and pose, as well as to estimate camera parameters. This is useful for applications such as virtual modelling of clothes in the fashion industry. A recent method reported by \citet{Mescheder:Occupancy:2019} is able to generate a realistic 3D surface from a single image intruding the idea of a continuous decision boundary within the deep neural network classifier. For 2D image to 3D object generation, generative models offer the best performance to date, with the state-of-the-art method, GAL \citep{Jiang:GAL:2018}, achieving an average IoU of 0.71 on the ShapeNet dataset. The creation of a 3D photograph from 2D images is also possible via tools such as SketchUp\footnote{https://www.sketchup.com/plans-and-pricing/sketchup-free} and Smoothie-3D\footnote{https://smoothie-3d.com/}. Very recently (Feb 2020), Facebook allowed users to add a 3D effect to all 2D images\footnote{https://ai.facebook.com/blog/powered-by-ai-turning-any-2d-photo-into-3d-using-convolutional-neural-nets/}. They trained a CNN on millions of pairs of public 3D images with their associating depth maps. Their Mesh R-CNN \citep{Gkioxari:mesh:2019} leverages the Mask R-CNN approach \citep{He:Mask:2017} for object recognition and segmentation to help estimate depth cues. A common limitation when converting a single 2D image to a 3D representation is associated with occluded areas that require spatial interpolation.

AI has also been used to increase the dimensionality of audio signals. Humans have  an ability to spatially locate a sound as our brain can sense the differences between arrival times of sounds at the left and the right ears, and between the volumes (interaural level) that the left and the right ears hear. Moreover, our ear flaps distort the sound telling us whether the sound emanates in front of or behind the head. With this knowledge, \citet{Gao:25Dsound:2019} created binaural audio from a mono signal driven by  the subject's visual environment to enrich the perceptual experience of the scene. This framework exploits U-Net to extract audio features, merged with visual features extracted from ResNet to predict the sound for the left and the right channels. Subjective tests indicate that this method can improve realism and the sensation being in a 3D space. { 
\citet{Morgado:self:2018} expand mono audio, recorded using a 360$^\circ$ video camera, to the sound over the full viewing surface of sphere. The process extracts semantic environments from the video with CNNs and  then high-level features of vision and audio are combined to generate the sound corresponding to different viewpoints. \citet{Vasudevan:Semantic:2020} also include depth estimation to improve realistic quality of super-resolution sound.
}

\subsection{Data Compression}

Visual information is the primary consumer of communications bandwidth across broadcasting and internet communications. The demand for increased qualities and quantities of visual content is particularly driven by the creative media sector, with increased numbers of users expecting increased quality and new experiences. Cisco predict, in their Video Network Index report,  \citep{cisco:2018} that there will be 4.8 zettabytes (4.8$\times 10^{21}$ bytes) of global annual internet traffic by 2022 -- equivalent to all movies ever made crossing global IP networks in 53 seconds. Video will account for 82 percent of all internet traffic by 2022. This will be driven by increased demands for new formats and more immersive experiences with multiple viewpoints, greater interactivity, higher spatial resolutions, frame rates and dynamic range and wider color gamut. This is creating a major tension between available network capacity and required video bit rate. Network operators, content creators and service providers all need to transmit the highest quality video at the lowest bit rate and this can only be achieved through the exploitation of content awareness and perceptual redundancy to enable better video compression.
 
Traditional image encoding systems (e.g., JPEG) encode a picture without reference to any other frames. This is normally achieved by exploiting spatial redundancy through transform-based decorrelation followed by variable length, quantization and symbol encoding.  While video can also be encoded as a series of still images, significantly higher coding gains can be achieved if temporal redundancies are also exploited. This is achieved using inter-frame motion prediction and compensation. In this case the encoder processes the low energy residual signal remaining after prediction, rather than the original frame. A thorough coverage of image and video compression methods is provided by \citet{David:Intelligent:2021}.

Deep neural networks have gained popularity for image and video compression in recent years and can achieve consistently greater coding gain than conventional approaches.  Deep compression methods are also now starting to be considered as components in mainstream video coding standards such as VVC and AV2. They have been applied to optimize a range of coding tools including intra prediction \citep{Schiopu:CNN:2019, Li:Fully:2018}, motion estimation \citep{Zhao:EnhancedM:2019}, transforms \citep{Liu:CNN:2018}, quantization \citep{Liu:One:2019}, entropy coding \citep{Zhao:Enhanced:2019} and loop filtering \citep{Lu:DVC:2019}. Post processing is also commonly applied at the video decoder to reduce various coding artefacts and enhance the visual quality of the reconstructed frames (e.g., \citep{Zhang:Enhancing:2020, Xue:Attention:2019}). Other work has implemented a complete coding framework based on neural networks using end-to-end training and optimisation \citep {Lu:end:2020}. This approach presents a radical departure from conventional coding strategies and, while it is not yet competitive with state-of-the-art conventional video codecs, it holds significant promise for the future.

 Perceptually based resampling methods based on {SR} methods using CNNs and GANs have been introduced recently.  Disney Research proposed a deep generative video compression system \citep{Han:Deep:2019} that involves downscaling using a VAE and entropy coding via a deep sequential model.  ViSTRA2 \citep{Zhang:ViSTRA2:2019}, exploits adaptation of spatial resolution and effective bit depth, downsampling these parameters at the encoder based on perceptual criteria, and up-sampling at the decoder using a deep convolutional neural network. ViSTRA2 has been integrated with the reference software of both the HEVC (HM 16.20) and VVC (VTM 4.01), and evaluated under the Joint Video Exploration Team Common Test Conditions using the Random Access configuration. Results show consistent and significant compression gains against HM and VVC based on Bj{\o}negaard Delta measurements, with average BD-rate savings of 12.6\% (PSNR) and 19.5\% (VMAF) over HM and 5.5\% and 8.6\% over VTM. This work has been extended to a GAN architecture by \citet{Ma:GAN:2020}. {Recently, \citet{Mentzer:High:2020} optimize a neural compression scheme with a GAN, yielding reconstructions with high perceptual fidelity. \citet{Ma:CVEGAN:2021} combine several quantitative losses to achieve maximal perceptual video quality when training a relativistic sphere GAN.} 
 
Like all deep learning applications, training data is a key factor in compression performance. Research by  \citet{Ma:BVIDVC:2020} has demonstrated the importance of large and diverse datasets when developing CNN-based coding tools. Their BVI-DVC database is publicly available and produces significant improvements in coding gain across a wide range of deep learning networks for coding tools such as loop filtering and post-decoder enhancement.
An extensive review of AI for compression can be found in \citep{David:Intelligent:2021} and  \citep{Ma:Image:2020}.





\section{Future Challenges for AI in the Creative Sector}
\label{sec:discussion}

There will always be philosophical and ethical  questions relating to the creative capacity, ideas and thought processes, particularly where computers or AI are involved. The debate often focuses on the fundamental difference between humans and machines. In this section we will briefly explore some of these issues and comment on their relevance to and impact on the use of AI in the creative sector.

\subsection{Ethical Issues, Fakes and Bias}

An AI-based machine can work `intelligently', providing an impression of understanding but nonetheless performing without `awareness' of wider context. It can however offer probabilities or predictions of what could happen in the future from several candidates, based on the trained model from an available database.  With current technology, AI cannot truly offer broad context, emotion or social relationship. However, it can affect modern human life culturally and societally. UNESCO has specifically commented on the potential impact of  AI  on culture, education, scientific knowledge, communication and information provision particularly relating to the  problems of the digital divide\footnote{ https://ircai.org/project/preliminary-study-on-the-ethics-of-ai/}. AI seems to amplify the gap between those who can and those who cannot use new digital technologies, leading to increasing inequality of information access. In the context of the creative industries, UNESCO mentions that collaboration between intelligent algorithms and human creativity may eventually bring important challenges for the rights of artists. 

One would expect that the authorship of AI creations resides with those  who develop the algorithms that drive  the art work. Issues of piracy and originality thus need special attention and careful definition, and deliberate and perhaps unintentional exploitation needs to be addressed. We must be cognizant of how easy AI technologies can be accessed and used in the wrong hands. AI systems are now becoming very competent at creating fake images, videos, conversations, and all manner of content. Against this, as reported in Section \ref{deepfakes}, there are also other AI-based methods under development that can, with some success, detect these fakes. 

The primary learning algorithms for AI are data-driven. This means that, if  the data used for training are unevenly distributed or unrepresentative due to  human selection criteria or labeling, the results after learning can equally be biased and ultimately judgemental. For example, streaming media services suggest movies that the users may enjoy and these suggestions must not privilege specific works over others. Similarly face recognition or autofocus methods must be trained on a broad range of skin types and facial features to avoid failure for certain ethnic groups or genders. Bias in algorithmic decision-making is also a concern of governments across the world\footnote{https://www.gov.uk/government/publications/interim-reports-from-the-centre-for-data-ethics-and-innovation/interim-report-review-into-bias-in-algorithmic-decision-making}. Well-designed  AI systems can not only increase the speed and accuracy with which decisions are made, but they can also  reduce human bias in decision-making processes. However, throughout the lifetime of a trained AI system, the complexity of data it processes is likely to grow, so even a network originally trained with balanced data may consequently establish some bias. Periodic retraining may therefore be needed. A review of various sources of bias in ML is provided in \citep{Ntoutsi:Bias:2020}.

\citet{Dignum:Ethics:2018} provide a useful classification of the relationships between ethics and AI, defining three categories: i) Ethics by Design, methods that ensure ethical behaviour in autonomous systems,  ii) Ethics in Design, methods that support the analysis of the ethical implications of AI systems, and iii) Ethics for Design, codes and protocols to ensure the integrity of developers and users. A discussion of ethics associated with AI in general can be found in \citep{Bostrom:ethics:2014}.

AI can, of course, also be used to help identify and resolve ethical issues.  For example, Instagram uses an anti-bullying AI\footnote{https://about.instagram.com/blog/announcements/instagrams-commitment-to-lead-fight-against-online-bullying} to identify negative comments before they are published and asks users to confirm if they really want to post such messages.

\subsection{The human in the Loop -- AI and Creativity}

Throughout this review we have recognized and reported on the successes of AI in supporting and enhancing processes within constrained domains where there is good availability of data as a basis for ML. We have seen that AI-based techniques work very well when they are used as tools for information extraction, analysis and enhancement. Deep learning methods that {characterize} data from low-level features and connect these to extract semantic meaning are well suited to these applications.  AI can thus be used with success, to perform tasks that are too difficult for humans or are too time-consuming, such as searching through a large database and examining its data to draw conclusions. Post production workflows will therefore see increased use of AI, including enhanced tools for denoising, colorization, segmentation, rendering and tracking. Motion and volumetric capture methods will benefit from enhanced parameter selection and rendering tools. Virtual production methods and games technologies will see greater convergence and  increased reliance on AI methodologies. 

In all the above examples, AI tools will not be used in isolation as a simple black box solution. Instead, they must be designed as part of the associated workflow and incorporate a feedback framework with the human in the loop. For the foreseeable future, humans will need to check the outputs from AI systems, make critical decisions, and feedback `faults' that will be used to adjust the model. In addition, the interactions between audiences or users and machines are likely to become increasingly common.  For example, AI could help to create characters that learn context in location-based storytelling and begin to understand the audience and adapt according to interactions.

Currently, the most effective AI algorithms still rely on supervised learning, where ground truth data readily {exist} or where humans have labeled the dataset prior to using it for training the model (as described in Section \ref{ssec:datasets}).   In contrast, truly creative processes do not have pre-defined outcomes that can simply be classed as good or bad. Although many may follow contemporary trends or be in some way derivative, based on known audience preferences, there is no obvious way of measuring the quality of the result in advance. Creativity almost always involves combining ideas, often in an abstract yet coherent way, from different domains or multiple experiences, driven by curiosity and experimentation. Hence, labeling of data for  these applications is not straightforward or even possible in many cases. This leads to difficulties in using current ML technologies. 

In the context of creating a new artwork, generating low-level features from semantics is a one-to-many relationship, leading to inconsistencies between outputs. For example, when asking a group of artists to draw a cat, the results will all differ in color, shape, size, context and pose. Results of the creative process are thus unlikely to be structured, and hence may not be suitable for use with ML methods. We have previously referred to the potential of generative models, such as GANs, in this respect, but these are not yet sufficiently robust to consistently create results that are realistic or valuable. Also, most GAN-based methods are currently limited to the generation of relatively small images and are prone to artefacts at transitions between foreground and background content. It is clear that significant additional work is needed to extract significant value from AI in this area.

\subsection{The future of AI technologies}

Research into, and development of, AI-based solutions continue apace. AI is attracting major investments from governments and large international organisations alongside venture capital investments in start-up enterprises. ML algorithms will be the primary driver for most AI systems in the future and AI solutions will, in turn, impact an even wider range of sectors. The pace of AI research has been predicated, not just on innovative algorithms (the basics are not too dissimilar to those published in the 1980s), but also on our ability to generate, access and store massive amounts of data, and on advances in graphics processing architectures and parallel hardware to process these massive amounts of data. New computational solutions such as quantum computing, will likely play an increasing role in this respect \citep{Welser:Future:2018}.

In order to produce an original work, such as music or abstract art, it would be beneficial to support increased diversity and context when training AI systems. The quality of the solution in such cases is difficult to define and will inevitably depend on audience preferences and popular contemporary trends.  High-dimensional datasets that can represent some of these characteristics will therefore be needed. Furthermore, the loss functions that drive the convergence of the network's internal weights must reflect perceptions rather than simple mathematical differences. Research into such loss functions that better reflect human perception of performance or quality is therefore an area for further research. 

 ML-based AI algorithms are data-driven; hence how to select and prepare data for creative applications will be key to future developments. Defining, cleaning and {organizing} bias-free data for creative applications are not straightforward tasks.  Because the task of data collection and labeling can be highly resource intensive,labeling services are expected to become more popular in the future. Amazon currently offers a cloud management tool, SageMaker\footnote{https://docs.aws.amazon.com/sagemaker/latest/dg/sms.html},  that uses ML to determine which data in a dataset needs to be labeled by humans, and consequently sends this data to human annotators through its Mechanical Turk system or via third party vendors. This can reduce the resources needed by developers during the key data preparation process.  In this or other contexts, AI may converge with blockchain technologies.  Blockchains create decentralized, distributed, secure and transparent networks that can be accessed by anyone in public (or private) blockchain networks. Such systems may be a means of trading trusted AI assets, or alternatively AI agents may be trusted to trade other assets (e.g., financial (or creative) across blockchain networks.  Recently, Microsoft has tried to improve small ML models hosted on public blockchains and plan to expand to more complex models in the future\footnote{https://www.microsoft.com/en-us/research/blog/leveraging-blockchain-to-make-machine-learning-models-more-accessible/}. Blockchains {make} it possible to reward participants who help to improve models, while providing a level of trust and security.  

As the amount of unlabeled data grows dramatically, unsupervised or self-supervised ML algorithms are prime candidates for underpinning future advancements in the next generation of ML. There exist  techniques that employ neural networks to learn statistical distributions of input data and then transfer this to the distribution of the output data \citep{Zhu:Unpaired:2017,Damodaran:DeepJDOT:2018,Xu:Learning:2019}. These techniques do not require a precise matching pair between the input and the ground truth, reducing the limitations for a range of applications. 

It is clear that current AI methods do not mimic the human brain, or even parts of it, particularly closely. The data driven learning approach with error backpropagation is not apparent in human learning. Humans learn in complex ways that combine genetics, experience and prediction-failure reinforcement. A nice example is provided by Yan LeCun of NYU and Facebook\footnote{LeCun credits Emmanuel Dupoux for this example.} who describes a 4-6 month old baby being shown a picture of a toy floating in space; the baby shows little surprise that this object defies gravity. Showing the same image to the same child at around 9 months produces a very different result, despite the fact that it is very unlikely that the child has been explicitly trained about gravity. It has instead learnt by experience and is capable of transferring its knowledge across a wide range of scenarios never previously experienced. This form of reinforcement and transfer learning holds significant potential for the next generation of ML algorithms, providing much greater {generalization} and scope for innovation. 

Reinforcement Learning generally refers to a goal-oriented approach, which learns how to achieve a complex objective through reinforcement via penalties and rewards based on its decisions over time. Deep Reinforcement Learning (DRL) integrates this approach into a deep network which, with little initialisation and through self-supervision, can achieve extraordinary performance in certain domains. Rather than depend on manual labeling, DRL automatically extracts weak annotation information from the input data, reinforced over several steps. It thus learns the semantic features of the data, which can be transferred to other tasks. DRL algorithms can beat human experts playing video games and the world champions of Go. The state of the art in this area is progressing rapidly and the potential for strong AI, even with ambiguous data in the creative sector is significant. However, this will require major research effort as the human processes that underpin this are not well understood. 

\section{Concluding Remarks}
\label{sec:conclusion}

This paper has presented a comprehensive review of current AI technologies and their applications, specifically in the context of the creative industries.  We have seen that ML-based AI has advanced the state of the art across a range of creative applications including content creation, information analysis, content enhancement, information extraction, information enhancement and data compression.  ML--AI methods are data driven and benefit from recent advances in computational hardware and the availability of huge amounts of data for training -- particularly image and video data. 

We have differentiated throughout between the use of ML--AI as a creative tool and its potential as a creator in its own right.    We foresee, in the near future, that AI will be adopted much more widely as a tool or collaborative assistant for creativity, supporting acquisition, production, post-production, delivery and interactivity. The concurrent advances in computing power, storage capacities and communication technologies (such as 5G)  will support the embedding of AI processing within and at the edge of the network. In contrast, we observe that, despite recent advances,  significant challenges remain for AI as the sole generator of original work. ML--AI works well when there are clearly defined problems that do not depend on external context or require long chains of inference or reasoning in decision making. It also benefits significantly from large amounts of diverse and unbiased data for training. Hence, the likelihood  of AI (or its developers)  winning awards for creative works in competition with human creatives may be some way off.  We therefore conclude that, for creative applications, technological developments  will, for some time yet, remain human-centric -- designed to augment, rather than replace, human creativity. As AI methods begin to pervade the creative sector, developers and deployers must however continue to build trust; technological advances must go hand-in-hand with a greater understanding of ethical issues, data bias and wider social impact.



\begin{acknowledgements}
This work has been funded by Bristol+Bath Creative R+D under AHRC grant AH/S002936/1. The Creative Industries Clusters Programme is managed by the Arts \& Humanities Research Council as part of the Industrial Strategy Challenge Fund.

The authors would like to acknowledge the following people who provided valuable contributions that enabled us to improve the quality and accuracy of this review: Ben Trewhella (Opposable Games), Darren Cosker (University of Bath), Fan Zhang (University of Bristol), and Paul Hill (University of Bristol).
\end{acknowledgements}

\bibliographystyle{spbasic}
\bibliography{literature_review}

\end{document}